%% file: neurips_2023.tex
\documentclass{article}

\usepackage[preprint, nonatbib]{neurips_2023}

\usepackage[utf8]{inputenc} 
\usepackage[T1]{fontenc}    
\usepackage{hyperref}       
\usepackage{url}            
\usepackage{booktabs}       
\usepackage{amsfonts}       
\usepackage{nicefrac}       
\usepackage{microtype}      
\usepackage{xcolor}         
\usepackage{graphicx}
\usepackage{multirow}

\input{content/00macro}

\title{Leveraging Large Language Model for Automatic Evolving of Industrial Data-Centric R\&D Cycle}

\author{%
    \textbf{Xu Yang\textsuperscript{1,*}, Xiao Yang\textsuperscript{1,*}, Weiqing Liu\textsuperscript{1}, Jinhui Li\textsuperscript{1}} \\
  \textbf{Peng Yu\textsuperscript{1}, Zeqi Ye\textsuperscript{1}, Jiang Bian\textsuperscript{1}} \\
  \textsuperscript{1}Microsoft Research Asia \\
  \texttt{\{xuyang1,Xiao.Yang,Weiqing.Liu,v-jinhuili,v-ypen,v-zeqiye,Jiang.Bian\}} \\
  \texttt{@microsoft.com} \\
  \textsuperscript{*}These authors contributed equally. \\
}

\begin{document}

\maketitle

\begin{abstract}
In the wake of relentless digital transformation, data-driven solutions are emerging as powerful tools to address multifarious industrial tasks such as forecasting, anomaly detection, planning, and even complex decision-making. Although data-centric R\&D has been pivotal in harnessing these solutions, it often comes with significant costs in terms of human, computational, and time resources. This paper delves into the potential of large language models (LLMs) to expedite the evolution cycle of data-centric R\&D. Assessing the foundational elements of data-centric R\&D, including heterogeneous task-related data, multi-facet domain knowledge, and diverse computing-functional tools, we explore how well LLMs can understand domain-specific requirements, generate professional ideas, utilize domain-specific tools to conduct experiments, interpret results, and incorporate knowledge from past endeavors to tackle new challenges.
We take quantitative investment research as a typical example of industrial data-centric R\&D scenario and verified our proposed framework upon our full-stack open-sourced quantitative research platform Qlib and obtained promising results which shed light on our vision of automatic evolving of industrial data-centric R\&D cycle.
\end{abstract}

\section{Introduction}
\label{section:intro}
Large language models (LLMs) are neural network-based systems that can generate natural language texts based on various inputs, such as keywords, prompts, or queries. LLMs have shown remarkable capabilities in various natural language processing (NLP) tasks, such as text summarization, machine translation, question answering, and text generation. However, their applications in other domains, such as research and development (R\&D), are still largely unexplored.

R\&D is the process of creating new or improving existing products or services. R\&D is vital for companies to maintain their competitiveness and innovation in many industries. We focus on a special subset of industrial R\&D scenarios, which we call \emph{data-centric R\&D scenarios}. These are the scenarios that are most likely to benefit from LLMs first. A data-centric R\&D scenario is one where data is the primary source of innovation and decision making in R\&D cycle. These scenarios share three common points: 1) A large amount of scenario \textbf{data} have been collected and organized, around which most of the R\&D activities centre. 2) The processing and mining of these data highly depend on domain \textbf{knowledge}. 3) There often exist domain-specific professional \textbf{tools} for conducting experiments on the data as well as analyzing the results afterwards.

\begin{figure}
  \centering
  \includegraphics[width=0.7\linewidth]{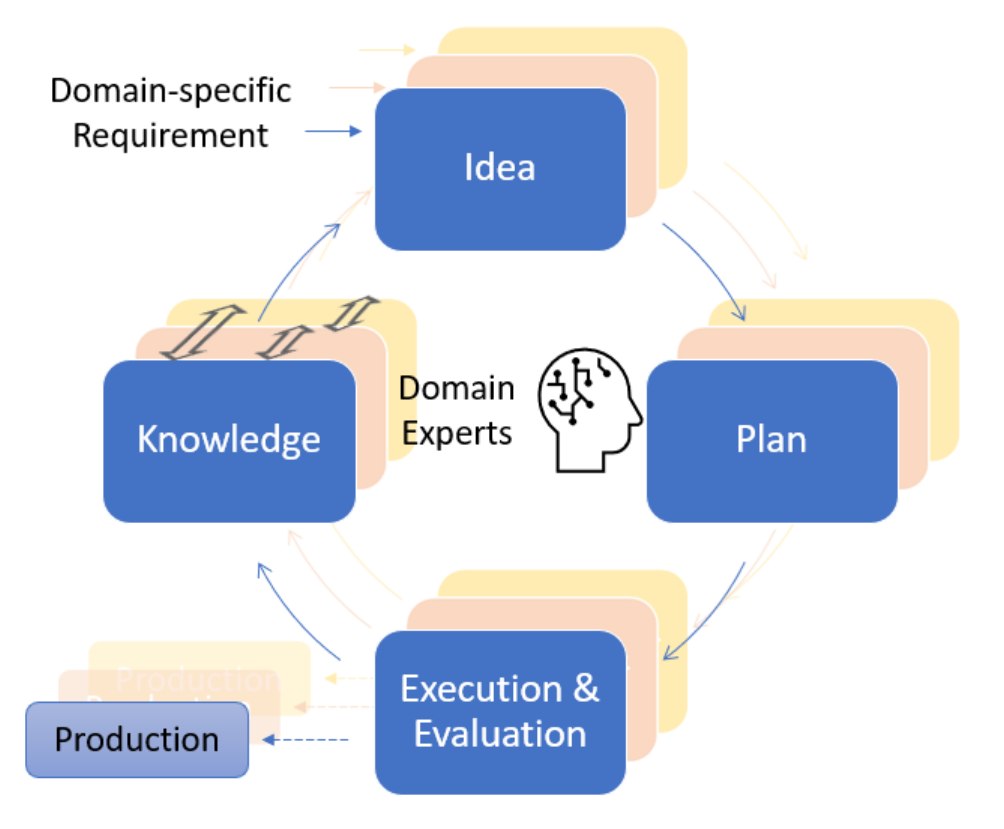}
  \caption{Illustration of R\&D evolving cycle.}
  \label{R&D loop}
\end{figure}

From above points, it is easy to understand that why data-centric R\&D requires highly skilled domain experts. Figure \ref{R&D loop} depicts a conceptual cycle of how domain-specific requirements are transformed into ideas and executable plans, and how domain knowledge is updated incrementally by analyzing the results, leading to further and better ideas. However, the traditional data-centric R\&D activities are almost exclusively driven by a very limited number of domain-specific experts, which results in high costs and long time frames for the R\&D evolving cycle. Moreover, the cutting-edge experiences and knowledge gained from one expert’s experiments on a specific R\&D task are hardly transferable to other experts and other related tasks, even within the same company or community.

In this paper, we propose to use LLMs as a tool to assist R\&D activities.  We hypothesize that LLMs can help reduce the human effort and enhance the quality of R\&D outputs by providing automated support for idea generation, experiment planning, result analysis, and idea evolution. To test our hypothesis, we focus on the following research questions:

Based on all the provided \textbf{data}, \textbf{knowledge} and \textbf{tools} of a specific data-centric domain, 






RQ1: How well can LLMs understand the domain-specific R\&D requirements?

RQ2: How well can LLMs generate relevant and reasonable professional ideas that effectively exploit domain knowledge and historical experience, while also being worth exploring for gaining new information?

RQ3: How well can LLMs comprehend professional domain-specific tools and schedule experiments to test and evaluate the ideas?

RQ4: How well can LLMs leverage the knowledge and experiences gained from previous R\&D activities to meet new requirements?



We choose machine learning-based quantitative investment as a typical example of data-centric R\&D. Quantitative investment is a data-driven approach to financial market analysis and decision making that relies on mathematical models and algorithms. Machine learning is a branch of artificial intelligence that enables computers to learn from data without explicit programming. Machine learning-based quantitative investment combines these two fields to develop intelligent systems that can automatically discover patterns, generate insights, and make predictions from financial data, which require strong explainability and heavily depend on domain knowledge and professional tools.

More specifically, we use Qlib, an open-sourced AI-oriented quantitative investment platform, as the domain-specific professional tool. Qlib provides a comprehensive set of tools for data processing, feature engineering, model training, backtesting, evaluation, and online deployment of machine learning-based quantitative investment strategies. Qlib also offers a rich collection of datasets covering various financial markets around the world.

Our research aims to answer four research questions by following a five-step process: 1) For each research question, we will construct a set of scenario-specific examples and corresponding evaluation metrics as the benchmark. 2) We will then build several existing and straight-forward LLM-based solutions and test and evaluate their performance on these tasks. 3) We will also analyze the challenges and obstacles of applying LLMs for these tasks in practice. 4) We will propose our paradigm design to address these challenges and obstacles and try to bridge the gap between potential and practice. 5) We will integrate our design for each research question into a full paradigm of LLMs enhanced R\&D evolving cycle and demonstrate its effectiveness in quantitative research.

Specifically, we encounter two major challenges when applying LLM to automate the R\&D cycle.
First, the R\&D process is complex, involving long-horizon planning and intricate actions. Some existing issues of LLMs such as hallucination \cite{ji2023survey}, may result in confidently outputting false facts or missing important steps.
We propose a methodology called "Framework as an Extensible Task-Dependent Symbolic Language" for robust and reliable planning.
Secondly, previous knowledge designs, also referred to as memory designs, are designed for general autonomous agents and do not take into account the specific requirements and characteristics of R\&D scenarios.
We have developed a dedicated knowledge management system that supports specific R\&D queries related to ideas, implementation, and tools.
In addition to applying LLMs to automate R\&D, we incorporate these R\&D-specific designs. The overall solution can be seen as a general paradigm for automating the industrial R\&D cycle. We refer to the proposed diagram as \textbf{\ppname{}} - an \textbf{A}utonomous \textbf{R\&D} \textbf{A}gent.

To summarize, our work has several contributions:
\begin{itemize}
    \item We are the first to formally define the R\&D evolving cycle and apply LLM to automate industrial R\&D.
    \item We identify the main challenges in applying LLM to automate R\&D, which include long-horizon planning and specific knowledge requirements. We propose solutions to address these challenges accordingly.
    \item We conduct extensive experiments on a concrete industrial scenario in quantitative investment and demonstrate the effectiveness of our work.
\end{itemize}

The rest of this paper is organized as follows: Section \ref{section:related} reviews the related work on LLMs and their applications in different domains. Section \ref{section:prel} formally defines the problem we are focusing on and provides preliminaries.  Section \ref{section:method} describes the methodology of our experiments. 
Section \ref{section:exp} presents the results and analysis of our experiments.
Section \ref{section:lim} discusses the limitations of our work.
Section \ref{section:concl} concludes the paper and suggests future work.

\input{content/02relatedworks}

\input{content/03preliminaries}
\input{content/04method}

\input{content/05experiment}

\input{content/06limdis}

\input{content/07conclu}


\medskip

\bibliography{ref} 

{
\small


\newpage

\appendix
\input{content/09suppliment}

\end{document}

%% file: content/00macro.tex
\usepackage{amsmath}        
\usepackage{todonotes}      
\usepackage{subfig}
\usepackage{appendix}

\newcommand{\ppname}{Arda}

\makeatletter
\renewcommand{\@noticestring}{}
\makeatother

%% file: content/02relatedworks.tex
\section{Related Works}
\label{section:related}

\subsection{Large Language Models}
In recent years, large language models (LLMs) have been incredibly successful in demonstrating their potential to achieve human-like intelligence,  as evidenced by their achievements \cite{openai2023gpt4, radford2019language, brown2020language, anthropic2023model, touvron2023llama2}.
Owing to their extensive training on copious amounts of textual data, these models have acquired an extraordinary capacity for processing and generating natural language.
Research indicates that LLMs surpassing a specific scale exhibit emergent capabilities \cite{wei2022emergent} and demonstrate exceptional performance in applications such as chatbots, machine translation, and text summarization \cite{zhao2023survey}.

The phenomenon of emergent abilities, which refers to capabilities that are absent in smaller models but become apparent in large models, is the most salient characteristic that sets Large Language Models (LLMs) apart from their predecessors.
Typical emergent abilities, such as in-context learning\cite{brown2020language}, instruction following\cite{ouyang2022training}, and step-by-step reasoning\cite{wei2022chain}, empower LLMs with the capabilities to handle diverse formats of problems in text.  

Automating the R\&D cycle involves many complex tasks that are highly flexible and require intelligence for intricate reasoning. 
Such tasks present considerable obstacles to achieving comprehensive automation.
LLMs have a significant impact on the AI community and prompt a rethinking of the possibilities of artificial general intelligence (AGI), as well as further automation of R\&D. 
Consequently, this paper delves into the development of advanced techniques inspired by the advancements in LLMs, aiming to address the complexities associated with automating R\&D processes.

\subsection{AutoML with LLMs}

AutoML \cite{hutter2019automated} refers to the process of automating the tasks of applying machine learning to real-world problems. 
This process incorporates every phase, starting with the acquisition of an unprocessed dataset and culminating in the creation of a deployable machine learning model.
AutoML aims to make machine learning available to non-machine learning experts, improve machine learning efficiency, and accelerate research on machine learning. Recently, some research works \cite{zhang2023mlcopilot, zheng2023can} aim to leverage LLMs to facilitate AutoML, improve explainability, and inject human knowledge.

Analogous to R\&D process, AutoML techniques necessitate a systematic exploration of optimal configurations through trial-and-error methodologies to effectively address various tasks.
Nevertheless, there are several key aspects in which AutoML essentially differs from the R\&D process.
Firstly, the R\&D process places significant emphasis on understanding domain-specific requirements, while AutoML predominantly focuses on well-defined problem settings without considering users' requirements. 
Secondly, the R\&D process is characterized by continuous evolution, wherein experiences and knowledge are updated to facilitate future tasks; however, AutoML tends to neglect the evolution of previous knowledge. 
Thirdly, the action space in the R\&D process is unstructured and highly complex, encompassing actions such as coding and configuring which are difficult to handle using structured parameter searching spaces in AutoML.
In conclusion, although there are similarities between the R\&D process and AutoML, the two are fundamentally distinct, with the R\&D process being a superset of AutoML. This distinction highlights the need for a comprehensive understanding of both methodologies to ensure their effective application in relevant contexts.

\subsection{Reinforcement learning with LLMs}
Automating the R\&D cycle necessitates a sequence of steps that entail engagement with the surrounding environment. This may encompass interactions with storage systems for the purpose of accessing and modifying code, as well as executing projects.    
Despite the similarity with with RL\cite{sutton2018reinforcement}, R\&D cycle are a more complicated process. 
Firstly, the action space in the R\&D cycle is frequently more complex than the structured action and state spaces found in RL. 
Secondly, R\&D typically employs fine-grained, decoupled components, such as those related to planning and execution, which adds to its complexity.

\subsection{Task automation with LLMs}

Task automation involves reducing human efforts by automatically completing specific tasks with autonomous agents\cite{zhang2023mlcopilot}. This has long been a prominent research topic in the academic community. Recently, LLMs have showcased remarkable intelligence and have been adopted to boost the capabilities of autonomous agents in various areas, such as planning\cite{yao2023tree} and tool utilization \cite{qin2023toollm}. 

LLMs provide exceptional intelligence. However, completing tasks requires a more complex process.
For instance, when an LLM is asked, "What should I do now to complete the task?" without any supplementary information, receiving a satisfactory response is unlikely. 
This is because there is a lack of current state information and action space, which are two key components in addition to \textbf{decision making}.
\begin{itemize}
    \item \textbf{Memory}: Language models are stateless and do not store information that is crucial for autonomous agents to complete dependent steps in a plan.
    \item \textbf{Grounding actions}: Grounding is a crucial aspect that informs LLMs about the available action space, enabling them to sense and interact with their external environment. By incorporating grounding, LLMs can execute external actions and process environmental feedback, subsequently integrating this information into their working memory as textual data.
\end{itemize}

\paragraph{General autonomous AI agent.}In recent years, there has been a growing number of autonomous AI agents for general purposes, such as those discussed in \cite{shen2023hugginggpt,yang2023auto}. These agents use general prompts to automatically plan, implement, and execute tasks with minimal assumptions about the scenario.
To improve task completion quality, we can make customized design decisions by leveraging scenario characteristics. Achieving such improvements in general autonomous AI agents is difficult.

\paragraph{R\&D autonomous agent.}
LLM provides significant benefits to many tasks, including scientific research and engineering. An LLM-based agent can serve as a research assistant in various aspects, from information management to proposing ideas and task planning \cite{boiko2023emergent,kang2023chatmof,bran2023chemcrow}.
They also demonstrate the ability to automate implementation tasks such as coding, testing, debugging, and documentation generation \cite{qian2023communicative,hong2023metagpt,dong2023self}. As a result, ideas can be implemented and tasks can be completed more efficiently.
However, existing works lack the ability to continuously research and develop. This involves the evolution of knowledge and solutions, which is a key part of the R\&D process. Automating this ability to continuously evolve is still largely unexplored.

%% file: content/03preliminaries.tex
\section{Preliminaries}

\label{section:prel}

In this section, we will formally define the R\&D process and the problem that we want to solve.

\subsection{A Formal Formulation of R\&D Process}
\begin{figure}
    \centering
    \includegraphics[width=1\linewidth]{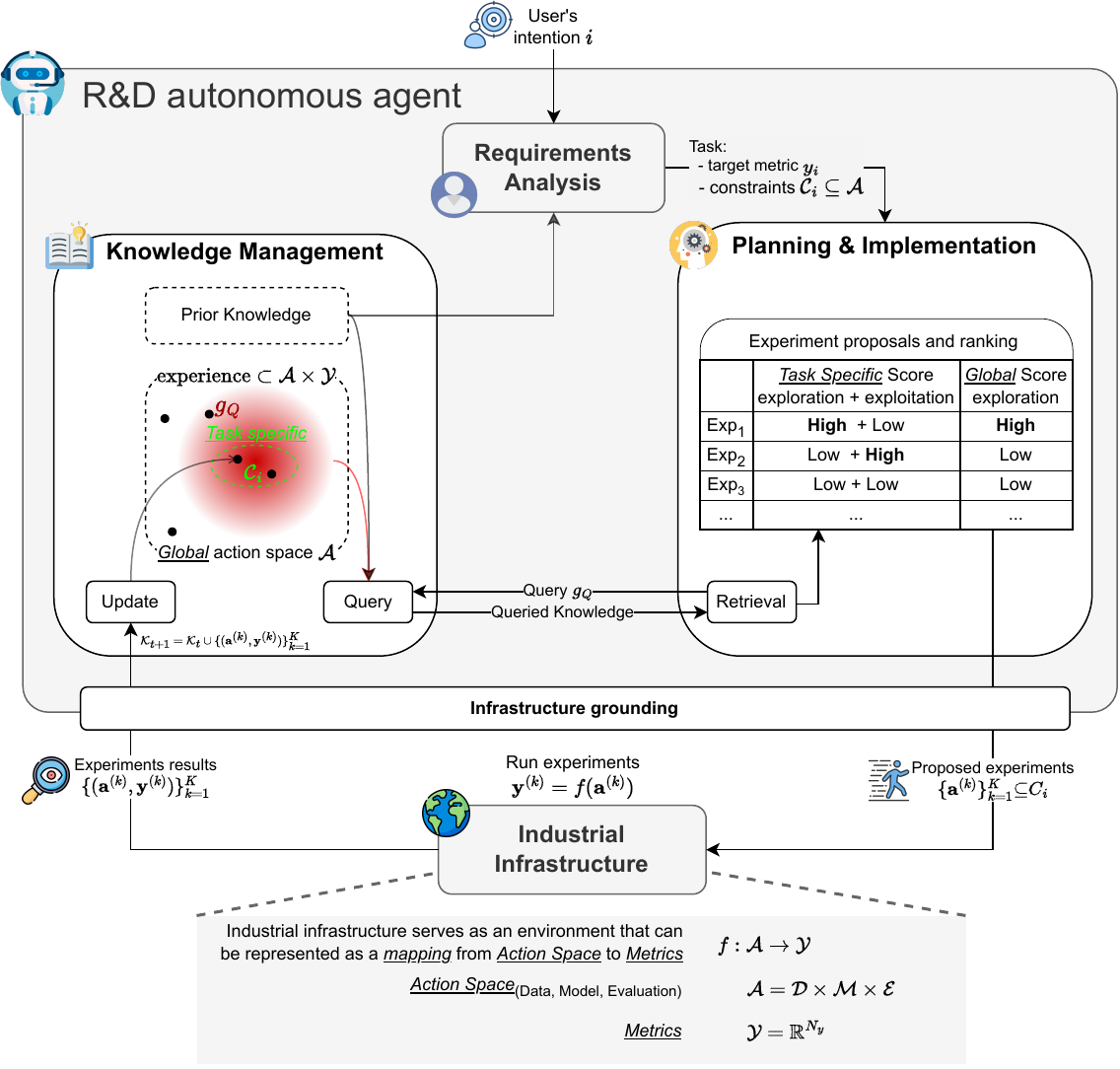}
    \caption{A formal formulation of R\&D process }
    \label{fig:formal}
\end{figure}

As shown in Figure \ref{fig:formal}, the R\&D process can be seen as the interactions between an agent and an environment. The agent proposes experiments, which can be considered as actions. These actions are implemented as runnable programs that are input into an industrial infrastructure to obtain experimental results as observations.
The agent will comprehend the user's intention and then iteratively collect information and exploit it to explore actions as well as possible under the users' requirements. Bayesian optimization \cite{snoek2012practical} is similar to the R\&D process and can be regarded as a special case of it. The R\&D process covers a more complex workflow and action space, which may involve writing code and configuration rather than searching within a regular hyperparameter space.

The action space can be denoted as $\mathcal{A} = \mathcal{D} \times \mathcal{M} \times \mathcal{E}$, where $\mathcal{D}$ represents an action space comprising a collection of functions. Each function corresponds to a specific data handling procedure, such as loading data from a file or normalizing. Likewise, $\mathcal{M}$ and $\mathcal{E}$ represent the action spaces of modeling and evaluation, respectively.
When an action $\mathbf{a} \in \mathcal{A}$ is selected and executed, the results can be evaluated using multiple metrics denoted by $\mathbf{y} \in \mathcal{Y}$ where $\mathcal{Y} = \mathbb{R}^{N_y}$.
The execution of an action takes place in industrial infrastructure, which can be represented as a mapping from the action space to the metrics space, denoted by $f: \mathcal A \rightarrow \mathcal Y$.
The R\&D process aims to explore the action space to find the best option that has the potential to improve the metric $y$, as per the user's requirements. Therefore, it is essential to have a rough estimation, which takes into account the expected and potential aspects, of $f$ for R\&D.

The action space is quite complex, involving comprehensive R\&D behaviors such as coding and configuring.
Designing the mapping function $f$ can be extremely difficult. Furthermore, each R\&D iteration typically requires significant effort, and obtaining data samples $\{(\mathbf a, \mathbf y)_1, (\mathbf a, \mathbf y)_2, \dots, (\mathbf a, \mathbf y)_{N_S}\}$ can be expensive. This results in sparse data samples, which further complicates the task.
Traditionally, the R\&D process is driven by domain experts.

\paragraph{Requirement analysis.}
The procedure commences with the user's intention $i$. Subsequently, a comprehensive requirements analysis is conducted to transform the user's intention into the precise target metric $y_i$ and the associated constraints $\mathcal{C}i$. 
It is essential to note that $\mathcal{C}{i} \subseteq \mathcal{A}$ serves to constrict the action space under examination. It is pertinent to consider all user preferences as constraints within this context. For instance, certain users may opt to maintain a fixed dataset and investigate the most suitable model that could potentially maximize performance enhancement.

\paragraph{Planning \& Implementation.}
After specifying the target variable $y_i$ and constraints $\mathcal C_i$, the next step is to select a promising research direction to explore.
In this phase,  knowledge is essential and can be categorized into prior knowledge and experiences.
Prior knowledge refers to general understanding and reasoning in a specific scenario. Experiences refer to previously executed $(\mathbf a, \mathbf y)$ experiences and are denoted as $\mathcal K_t = \{(\mathbf a, \mathbf y)_k\}_{k=1}^{N_K}$.
Experts evaluate the utilities of each research direction to propose a promising research direction, selecting the ones with the high utility.
A good research direction with high utility should consider both exploitation and exploration. Exploitation means that the proposed direction can averagely achieve a high value for $y_i$. Exploration means that the proposed direction has high uncertainty and therefore needs to be further explored.
In addition to considering exploitation and exploration in current tasks, some actions may provide value in future tasks.  
To effectively rank these actions, it is advisable to aggregate scores derived from various aspects. Consequently, experiments exhibiting high utility will be prioritized for proposal.
Drawing upon past experiences as exemplary cases, the agent can then efficiently propose well-informed experiments based on them.

\paragraph{Experiment execution and knowledge management.}
The final proposed experiments are $\{\mathbf a^{(k)}\}_{k=1}^{K}$.
Experiments results $\{(\mathbf a^{(k)}, \mathbf y^{(k)})\}_{k=1}^{K}$  will be produced after experiment execution $\mathbf y^{(k)} = f(\mathbf a^{(k)})$.    They will update knowledge base by $\mathcal K_{t+1}  = \mathcal K_t \cup \{(\mathbf a^{(k)}, \mathbf y^{(k)})\}_{k=1}^{K}$.  The updated knowledge base will provide experiences for future R\&D process. This type of knowledge is referred to as practice knowledge.

\subsection{Problem Formulation}

As stated in the previous section, the R\&D automation research primarily focuses on automating the R\&D activities mentioned above.
After formally formulating the R\&D process, we revisit previously proposed research questions and gain a deeper understanding of them.

\begin{itemize}
    \item  RQ1 evaluates if the parsed user requirements,  target metric $y_i$ and constraints $\mathcal C_i$, align with the original user intention $i$.
    \item  RQ2 evaluates the best experiment $\mathbf a_{\text{best}}$ for metric $y_i$ within a limited number of trials.
    \item  RQ3 evaluates the alignment between the proposed experiments $\{\mathbf a^{(k)}\}_{k=1}^{K}$ and the domain-specific tool.
    \item  RQ4 evaluates the best experiment $\mathbf a_{\text{best}}$ for metric $y_i$ within a limited number of trials while the knowledge comes from different requirements $y_i$ and constraints $\mathcal C_i$.
\end{itemize}

\subsection{LLM-Based R\&D Automation}

The daily R\&D process demands significant expertise and effort. While the high cost of this process creates an incentive to automate it, existing challenges make automation difficult.

\begin{itemize}
    \item The automation of the R\&D process calls for advanced artificial intelligence capabilities, which are essential for comprehending and reasoning within intricate domains.
    \item The action space is unstructured and extremely complicated. Some actions, such as coding and configuring, are difficult to be appropriately handled by models.
    \item Proposing a promising research direction to explore requires a good estimation of $f: \mathcal A \rightarrow \mathcal Y$. This estimation is complicated and usually relies on only a few examples. 
    As a result, the incorporation of a robust prior and proficiency in few-shot learning are crucial for obtaining precise estimations.
\end{itemize}

LLM has recently demonstrated remarkable potential in achieving expert-level intelligence in many fields. This brings hope for automating the R\&D process.

\begin{itemize}
    \item It demonstrates strong and growing intelligence and abilities in understanding and reasoning.
    \item It is capable of processing intricate information through languages, such as coding and configuring, further highlighting its versatility.
    \item Leveraging extensive amounts of training data, LLM has amassed a wealth of common-sense knowledge, which furnishes excellent priors. Its capacity to learn in context empowers it to manage sparse samples effectively.
\end{itemize}

LLM offers several advantages that make it an excellent solution for building an R\&D autonomous agent. This agent could potentially bring the R\&D process into a new era.

%% file: content/04method.tex
\section{Method}

\label{section:method}

This section introduces our method for implementing our R\&D autonomous agent.
Although LLM provides many features for automating R\&D, simply applying LLM to automate R\&D still faces challenges due to its unique characteristics, which are different from those of other autonomous agents.
The complexity of R\&D encompasses long-horizon planning and intricate actions, including configuration and coding. 
Moreover, continuous evolution serves as a crucial distinguishing factor for R\&D in comparison to alternative autonomous agents.

To automate the R\&D process:

\begin{itemize}

	\item We propose a methodology that uses an industrial framework as an extensible task-dependent Symbolic Language (SL) to support strong and robust long-horizon planning abilities. With this design, complex actions are decomposed into low-complexity, single-action tasks. These two advantages make long-horizon planning and implementation of complex actions in an industrial setting more feasible.

	\item To support continuous evolution, we propose a knowledge management method that facilitates knowledge updating and querying. This method enhances the performance of each step of the R\&D process.

\end{itemize}

In the following section, we will introduce the two parts in detail.

\subsection{Framework as an extensible task-dependent symbolic language}
\label{subsection:faaetsl}

\begin{figure}
    \centering
    \includegraphics[width=1\linewidth]{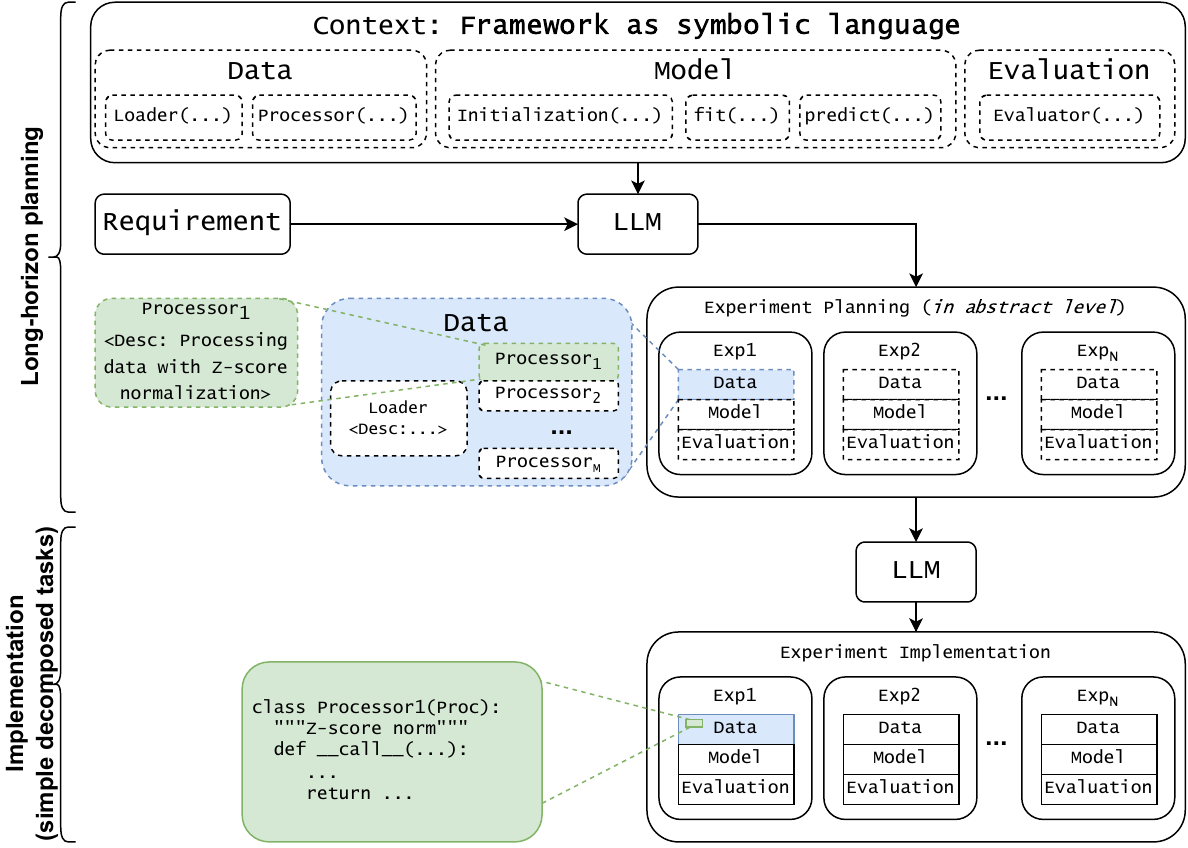}
    \caption{Automate R\&D by reformulating industrial framework as an extensible task-dependent symbolic language. The first step is to reformulate the industrial framework as an extensible, task-dependent symbolic language. This makes the planning process more robust and reliable. The decomposed tasks become simpler as a result, and can be easily implemented and extended by using LLM.}
    \label{fig:framework}
\end{figure}

The R\&D process is complex, involving long-horizon planning and intricate actions. LLM is still in rapid development and has some issues, such as hallucination \cite{ji2023survey}, which can output false facts confidently or miss important steps. This makes it difficult to complete R\&D tasks that require lengthy reasoning and intricate implementation steps.

\cite{liu2023llm} noticed such a challenge and found that classical planners, when given a problem in a formatted way, can use efficient search algorithms to quickly identify correct, or even optimal, plans. Firstly, the problem is converted to a standard Planning Domain Definition Language(PDDL) format, and then a symbolic planner is used to solve it. This solution performs exceptionally well in problems that are well-defined and easily structurally formalized.
However, many real-world scenarios are difficult to formalize in a structured way. This compromises the expressiveness of the solution. Planning involves a more complex process than structural planning.

\cite{lyu2023faithful, gao2023pal} propose a more flexible approach to support advanced planning abilities while maintaining process robustness and reliability.
They interleave natural language with a task-dependent Symbolic Language (SL), such as Python, Datalog, or PDDL, for planning. The LLM conducts planning based on the SL. The use of a standard task-dependent Symbolic Language (SL) helps to keep the process robust and reliable. A more comprehensive SL, combined with LLM's flexible and intelligent planning, provides a solution with wide expressiveness.

In this paper, we propose a methodology named "Framework as an Extensible Task-Dependent Symbolic Language."

Industrial solutions are characteristically intricate in nature. To simplify them, users often build solutions on a framework with high cohesion, loose coupling, and extensibility.
Such frameworks are frequently designed with modular components that are decoupled, allowing for versatile composition in order to develop complex solutions.

For robust and reliable planning, industrial frameworks can be a great task-dependent symbolic language.
The basic idea is similar to \cite{lyu2023faithful, gao2023pal}. We provide a novel perspective for automating R\&D planning by regarding the industrial framework as a symbolic language.
In addition, industrial frameworks, unlike other task-dependent symbolic languages, are often extensible. This provides enough expressiveness for the action space.
The extensible modules are often single-action tasks that are simple and well-aligned with LLM's area of expertise.
This type of solution can handle long-horizon planning without compromising the expressiveness of the action space, making it a great solution for R\&D automation.

Figure \ref{fig:framework} shows the framework of our proposed solution. The first step is to reformulate the industrial framework as an extensible, task-dependent symbolic language. This makes the planning process more robust and reliable. 
As a consequence, the decomposed tasks exhibit increased simplicity, thereby allowing for seamless implementation and extension through the utilization of LLMs.

\subsection{R\&D-specific knowledge management}

Knowledge management is an important aspect of our system design. 
This dynamic process perpetually evolves through the accumulation and updating of information, subsequently offering valuable insights via queries that have the potential to significantly inform and guide future actions.
Knowledge management can be seen as a form of memory management that plays a crucial role in the development of AI agents.
Previous works explore diverse designs of memory, including long-term and short-term memory \cite{shinn2023reflexion,lin2023swiftsage}, as well as embeddings \cite{lin2023swiftsage}, among others.
Nonetheless, these memory designs primarily target general autonomous agents and, as such, fail to address the unique requirements and characteristics inherent to R\&D scenarios.

To fully leverage the characteristics of R\&D automation and better meet the requirements, we have developed a dedicated knowledge management system.
R\&D automation encompasses multiple steps, each with its own sub-goals and information requirements for making informed decisions. 
The extensive nature of the knowledge base, which surpasses the restricted context length of LLM, necessitates the accurate querying of pertinent information. This is instrumental in aiding the LLM to make well-informed and appropriate decisions.

\begin{figure}
    \centering
    \includegraphics[width=0.5\linewidth]{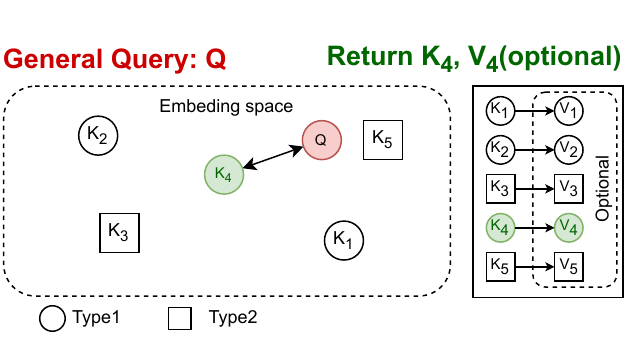}
    \caption{ General knowledge management based on embeddings.}
    \label{fig:GenralKM}
\end{figure}

In our system, we initially support general queries, as illustrated in Figure \ref{fig:GenralKM}.  Although we primarily focus on a special scenario (i.e., R\&D automation), certain requirements necessitate the utilization of general queries. For example, possessing domain knowledge may be useful when analyzing user's requirements.
As shown in Figure \ref{fig:GenralKM}, general knowledge is stored using a key-value format, where the value is optional. The keys are encoded in an embedding space with different categories. When a query $Q$ is received by the system with a specific type, it will return the most relevant knowledge.

\begin{figure}
\centering
\includegraphics[width=1.0\textwidth]{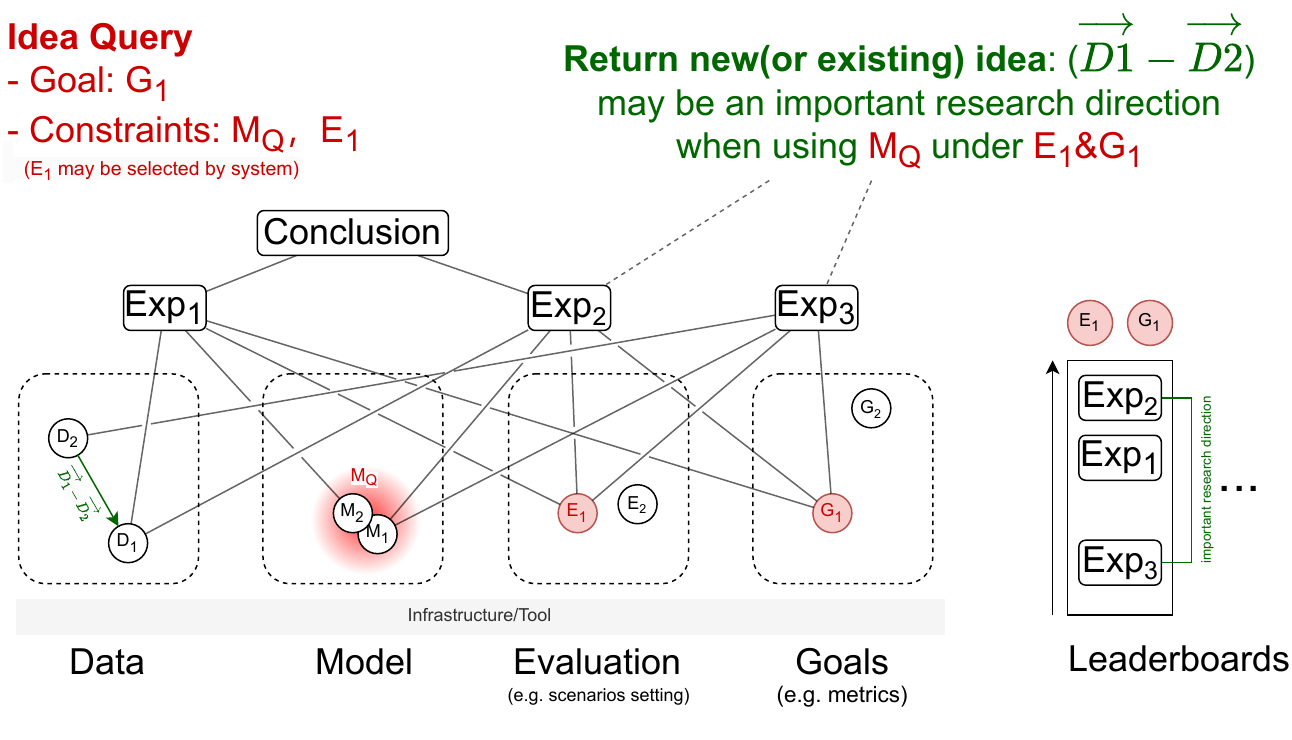}
\caption{Overall design of knowledge management and an example of idea query.}
\label{fig:KMAll}
\end{figure}






\begin{figure}
\centering
\subfloat[Implementation query.]{\includegraphics[width=0.43\textwidth]{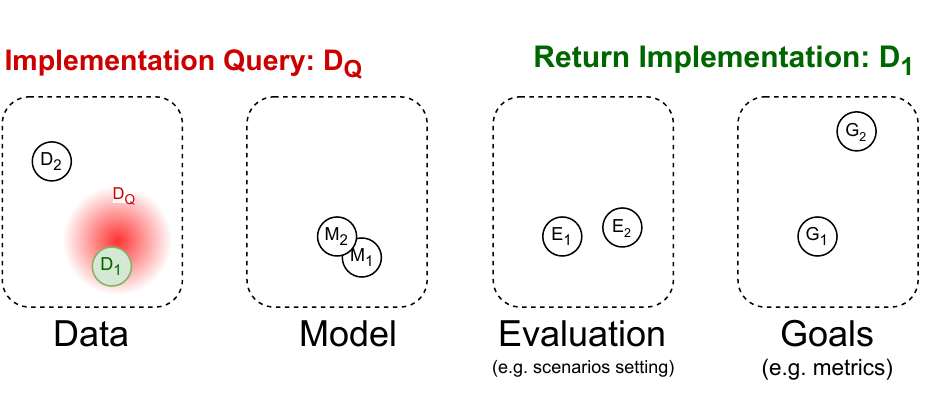}\label{fig:KMImp}}
\hfill 
\subfloat[Infrastructure query.]{\includegraphics[width=0.51\textwidth]{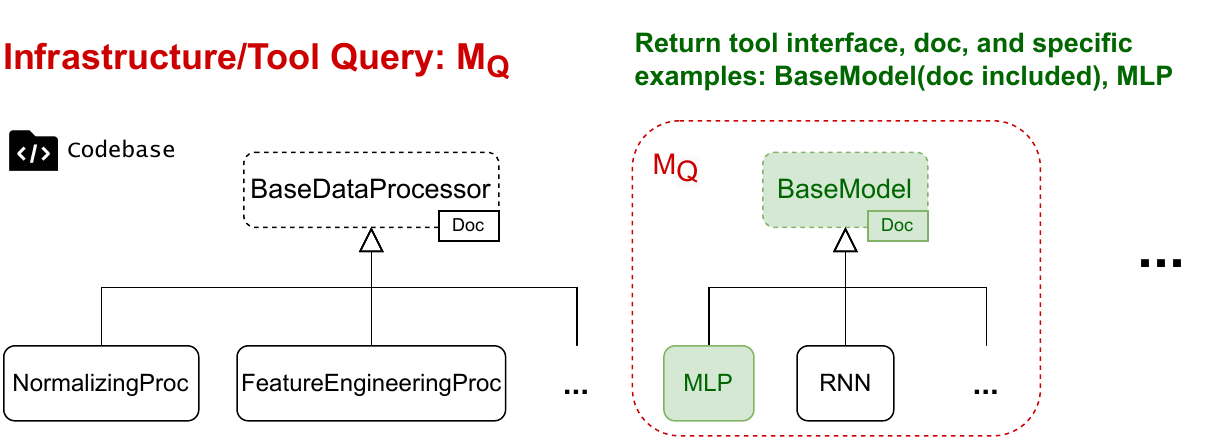}\label{fig:KMInfra}}
\caption{Extra features of knowledge management.}\label{fig:KMFea}
\end{figure}

In addition to supporting general requirements, the management of R\&D-specific knowledge is a crucial aspect of our design.
As illustrated in Figure \ref{fig:KMAll}, an experiment (referred to as a trial in R\&D) is comprised of several key components: data, model, evaluation, and goal. 
Each component is transformed into embeddings and stored in an embedding space. 
The distances between embeddings are expected to represent the differences in the specific component. 
A set of embeddings from different components is connected to an experiment. 
This indicates that the embeddings are extracted from the experiment. 
More detailed information is saved within the experiments. 
The primary purpose of the embeddings is to assist in querying.
Each experiment generates a set of results. Results obtained under the same evaluation settings and metrics can be compared on a leaderboard.

With this storage design, we will explain how it supports certain R\&D-specific queries.
Proposing a reasonable idea that meets the user's requirements is a crucial part of the R\&D process. Traditionally, this task is performed by experts. It involves leveraging previous knowledge to strike a balance between exploitation and exploration. Exploitation entails proposing ideas with high expected performance, while exploration involves actively exploring ideas with high potential (i.e., high variance). The system provides an idea query feature to support these requirements.

As shown in Figure \ref{fig:KMAll}, the user's intention is converted into a specific goal, $G_1$, and constraints, $M_Q$ and $E_1$. $G_1$ represents a specific metric. $M_Q$ may provide a description of the model, which could be relatively vague. $E_1$ refers to the evaluation setting.
The knowledge management base supports querying that returns previous relevant experiments that meet the given constraints. In this example, $\text{EXP}_2$ and $\text{EXP}_3$ are returned, and they differ greatly in leaderboards. Therefore, it may be worth exploring the difference, $\vec{D_1} - \vec{D_2}$, between $\text{EXP}_2$ and $\text{EXP}_3$ as a potential idea to consider. It is important to note that the datasets $M_1$ and $M_2$ are not exactly aligned, but they are similar. Therefore, inference on unseen conclusions is used to propose new ideas rather than existing ones.
In addition to evaluating variance to propose ideas worth exploring, exploitation can also be performed by proposing ideas under settings with higher average performance.

In addition to assisting in proposing ideas, we can query previous implementations as demonstrations. Providing the right demonstration is essential to improve the performance of LLM when translating plans into concrete runnable projects. As shown in Figure \ref{fig:KMImp}, previous implementations can serve as demonstrations for the next implementations. Implementations are often extensions of a framework. Therefore, understanding the interface and documentation of the framework is also essential information. As shown in Figure \ref{fig:KMInfra}, querying the codebase is supported.

%% file: content/05experiment.tex
\section{Experiment}

\label{section:exp}

In recent years, there has been a surge in research on LLM-based autonomous agents. Numerous datasets and benchmarks are available for evaluation \cite{chalamalasetti2023clembench, liu2023agentbench, qin2023tool, yao2022webshop, ziems2023can, lee2022evaluating, bran2023chemcrow, chang2023survey}. However, designing a general evaluation mechanism is challenging because agents are typically tailored to specific scenarios, which often have unique evaluation preferences different from general tasks.

We are the first to propose an autonomous agent for R\&D. However, benchmarks for assessing LLMs for R\&D autonomous agents are currently missing.
This presents a challenge in evaluating the approach we propose in this paper. The main challenges arise from the fact that it relies heavily on domain experts' understanding of the results, and not all aspects can be quantitatively evaluated.
To address this issue, we involve subjective evaluations from domain experts.

\subsection{Design of experiments}

The R\&D process is a complex workflow that involves many steps. To make the evaluation more comprehensive and informative, we conduct extensive experiments to separately evaluate the following key aspects of LLM-based R\&D autonomous agent, which corresponds to the research questions proposed in Section \ref{section:intro}.
\begin{itemize}
    \item Understanding: Can LLM accurately comprehend users' requirements?
    \item Exploration and exploitation: Can LLM exploit historical actions to propose professional ideas that have high potential value worth exploring?
    \item Grounding: Can LLM understand professional domain-specific tools and schedule experiments to test and evaluate ideas?
    \item Transferability: Can LLMs use the knowledge and experiences gained from previous R\&D activities to meet new requirements that may differ from the previous ones?
\end{itemize}

The upper aspects can be classified into two distinct scenarios: cold start and warm start. The primary distinction between these scenarios lies in the utilization of prior practice knowledge.
Understanding and Grounding are evaluated in cold start scenarios as they predominantly harness domain knowledge that does not accumulate over time. 
Conversely, Transferability is assessed in warm start scenarios due to its inherent reliance on the transfer of prior practice knowledge to new tasks.
Exploration and exploitation is a little bit tricky since it changes from cold start to warm start. 
At the inception of research, there is an absence of prior practice knowledge to exploit. As R\&D progresses, the exploitation of prior practice knowledge gains increasing significance.
Hence, we will present our experiment results in two tables, of which one is about the cold start while another one is in warm start scenario.
The metrics of tasks can be categorized into subjective domain-experts evaluation and objective metrics.
Now, let's introduce the overall design of each category of tasks.

For understanding, the compared methods take a user intention $i$ as input and generate requirement analysis results. These results consist of target requirements $y_i$ and constraints $\mathcal C_i$. The user's preferences will be considered as a type of constraint.Evaluating the analysis results quantitatively can be challenging. Therefore, we will seek assistance from domain experts to assign a score between 0-1 to the analysis results. User requirements are often only partially explicit, with some remaining implicit. The evaluation includes the ability to infer implicit requirements.

For exploration and exploitation, the compared methods will take specific requirements as input. These methods will then iteratively propose several experiments to determine the best solution within the users' constraints. Maximizing the effectiveness of the proposed experiments requires considering both exploration and exploitation. The methods should leverage existing knowledge to propose experiments in the action space with high expectations, while also exploring actions with high variance to uncover potentially better outcomes. 
In addition, we also seek domain expert's help to assign score between 0-1 to each group of experiments considering the accuracy, professionalism and feasibility of the experiments. Accuracy focuses on whether the experiments stick to user's intention. Professionalism focuses on whether the experiments are likely to the proposed by an expert. Feasibility focuses on how well our experiments align with our grounding skill. A weighted sum score will be presented to measure the results.

For grounding, a detailed plan will be taken as input and implemented as a runnable project. In order to quantitatively evaluate the grounding process, we will assess its success rate. However, the success rate alone cannot provide a comprehensive evaluation of the grounding performance and can be manipulated. For instance, compared methods may propose a runnable solution that accomplishes an easier task but does not align with users' requirements. Such a hack approach is difficult to automatically detect in a general manner. Therefore, domain experts will assist in evaluating the task to ensure a more comprehensive evaluation.

For transferability, the evaluation process is similar to that of exploration and exploitation. 
The difference is that it incorporates the experience gained from previous experiments conducted under different requirements. The knowledge acquired previously is expected to be applied to new tasks to enhance both exploitation and exploration.
domain expert is also asked to help evaluate the idea and experiments considering the exploitation of former practice knowledge.

To provide a clearer evaluation, we have created Table \ref{tab:evaldesign} to describe the evaluation design.

\begin{table*}[htbp]
\centering

\resizebox*{\columnwidth}{!}{%

\begin{tabular}{|l|l|l|l|}
\hline
Scenario & Phase & Evaluation & Number of cases \\ \hline
\multirow{3}{*}{Cold start} & Understanding & Requirements analysis & 10\\ \cline{2-4}
 & Exploration \& Exploitation & Accuracy \& Professionalism \& Feasibility & 10\\ \cline{2-4}
 & Grounding & Planning alignment \& Pass rate & 20\\ \hline
\multirow{2}{*}{Warm start} & Exploration \& Exploitation & Accuracy \& Professionalism \& Feasibility & 5\\ \cline{2-4}
 & Transferability & Exploitation and exploration of former tasks & 5\\ \hline
\end{tabular}
}

    \caption{The overall design of the evaluation.}
    \label{tab:evaldesign}
\end{table*}   

To evaluate these tasks, a specific R\&D scenario must be chosen. We have selected quantitative investment (Quant) as the scenario for our experiments.

Quant researchers iteratively optimize their models to better capture market patterns \cite{obthong2020survey}.
Like the typical workflow in R\&D, Quant often begins with specific user requirements, which may include desired metrics and evaluation settings. Researchers then propose ideas that involve datasets and models based on these requirements. The models often make predictions about key variables.
The next step is to evaluate the performance by constructing trading strategies based on these key variables and running backtests. Feedback is collected and used to propose better ideas in the next iteration.

\subsection{Compared methods}

We compare our methods with two baseline approaches.
\begin{itemize}
    \item \textbf{Standard} few-shot prompting\cite{brown2020language}: We utilize the in-context learning capability of LLM by providing demonstrations with input and output.
    \item \textbf{CoT} \cite{wei2022chain}: In addition to the Standard approach, we prompt LLM to output a reasoning chain before outputting the final answer.
\end{itemize}

%

For R\&D automation, the reasoning chain becomes more complex and requires long-horizon planning. We conducted experiments with general autonomous agents, like AutoGPT, but found it difficult to complete such a complex task without a design for grounding and planning. 
However, a failed baseline alone is not informative enough, so we attempted to propose a more informative baseline.
Therefore, methods are compared individually for each part instead of considering the end-to-end workflow.

Our method, \ppname{}, aims to improve the automation of the R\&D process by treating an industrial framework as a task-specific symbolic language, which enables more reliable planning. Qlib \cite{yang2020qlib} is a comprehensive Quant platform that serves as an excellent candidate for this task-specific symbolic language.  So we base \ppname{} on Qlib.

As described in Section \ref{section:method}, our method design mainly focuses on an industrial framework as a task-specific symbolic language for more reliable planning and domain-specific knowledge management. To better demonstrate the effectiveness of our design, the following methods are compared.

\begin{itemize}
    \item \textbf{\ppname{}} is a complete version of our method, an \textbf{A}utonomous \textbf{R\&D} \textbf{A}gent
    \item \textbf{\ppname{} wo KM} is \ppname{} without R\&D-specific knowledge management.
\end{itemize}


All of the methods mentioned above and their corresponding results were executed on GPT-4-32K in Azure OpenAI, using version 0613.

\subsection{Results}

As indicated in the Table \ref{tab:rescoldstart} and Table \ref{tab:reswarmstart}, our methods demonstrate superior performance compared to previous methods.

\begin{table*}[htbp]
\centering

\resizebox*{\columnwidth * 2 / 3}{!}{%
    \begin{tabular}{|l|l|l|l|l|}
    \hline
     & Understanding & Exploration \& Exploitation & Grounding & Overall \\ \hline
    Standard & 0.93& 0.73& 0.747 & 0.802 \\ \hline
    CoT & 0.93&0.74 & 0.825 & 0.831 \\ \hline
    \ppname{} & \textbf{0.99}& \textbf{0.917}& \textbf{0.995}& \textbf{0.967} \\ \hline
    \end{tabular}
}

    \caption{Evaluation results of different methods for each category of tasks on the first round of execution. All methods in this table have no practice knowledge when execution.}
    \label{tab:rescoldstart}

\end{table*}

\begin{table*}[htbp]
\centering

\resizebox*{\columnwidth * 2 / 3}{!}{%
    \begin{tabular}{|l|l|l|}
    \hline
      & Exploration \& Exploitation & Transferability \\ \hline
    Standard & 0.66& N/A  \\ \hline
    CoT & 0.685&N/A   \\ \hline
    \ppname{} wo KM & 0.82& 0.85 \\ \hline
    \ppname{} &\textbf{0.85}& \textbf{0.96} \\ \hline
    \end{tabular}
}

    \caption{Evaluation results of different methods for each category of tasks after several rounds of exploration. All methods can leverage the practice knowledge from the former exploration to help the following exploration.}
    \label{tab:reswarmstart}

\end{table*}  

As we can see in Table \ref{tab:rescoldstart}, the scores of each key evaluation aspect and the overall score in cold start scenario are displayed.
The Table \ref{tab:rescoldstart} shows the overall scores and scores of each key evaluation aspect in cold start scenario. The Table \ref{tab:reswarmstart} shows the scores in warm start.
For detailed scores, please refer to Appendix \ref{appdix_detail}.

In cold start scenario, overall, the Standard method performs the worst. The CoT method encourages the model to think step-by-step and achieve slightly better performance than the naive Standard method.
The gain in our method is derived from grounding and exploitation \& exploration, which necessitate design and implementation. As a result, they have a longer reasoning chain compared to other tasks. 
The performance gain in understanding is marginal because the understanding only involves one round of QA. The only difference comes from the queried knowledge for in-context learning, which provides very limited help. Since no abundant former practice knowledge is available in cold start scenario, \ppname{} wo KM performs exactly the same to \ppname{} so it is muted in Table \ref{tab:rescoldstart}.

As shown in table \ref{tab:reswarmstart}, in warm start scenario, Standard and COT mainly relies on the past conversation to maintain exploration which result in dissatisfaction. Transferability is unachievable since no memory is kept in Standard and COT. 
The performance gain in warm start scenario from our methods compared to baseline methods is more significant. This gain comes from two aspects. Firstly, \ppname{} wo KM and \ppname{} consider the framework as a symbolic language and use it during planning. This approach makes the planning and implementation more reliable and yields better results.
Secondly, retrieving the right knowledge at the right step is essential for making informed decisions. \ppname{} leverages R\&D-specific knowledge management and outperforms \ppname{} wo KM, which indicates the effectiveness of our knowledge management design.

\subsection{Case study}

\begin{figure}
\centering
\includegraphics[width=1.0\textwidth]{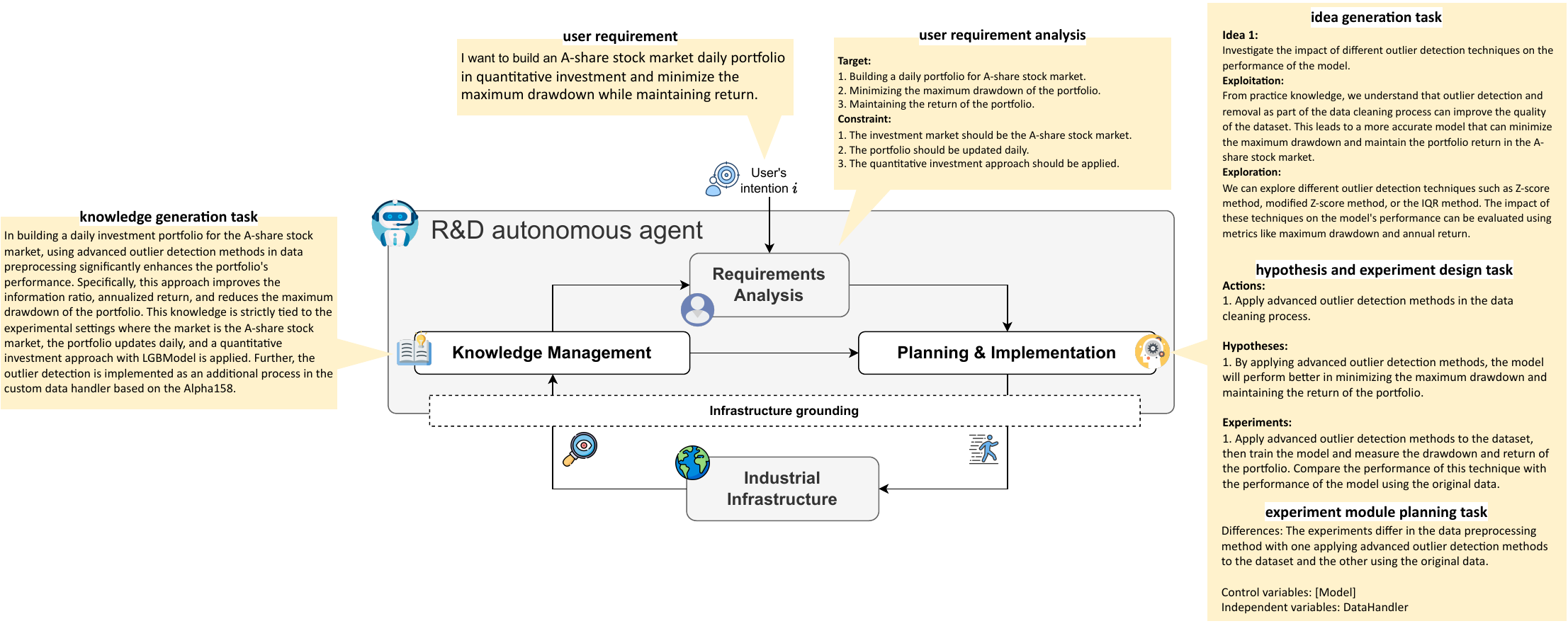}
\caption{Case study on requirement analysis, planning and knowledge management. content in the comment box are answers solely generated by \ppname{} in each sub-task. This case shows a research plan on outlier clip in dataset.}
\label{fig:case_study_plan}
\end{figure} 

\begin{figure}
\centering
\includegraphics[width=1.0\textwidth]{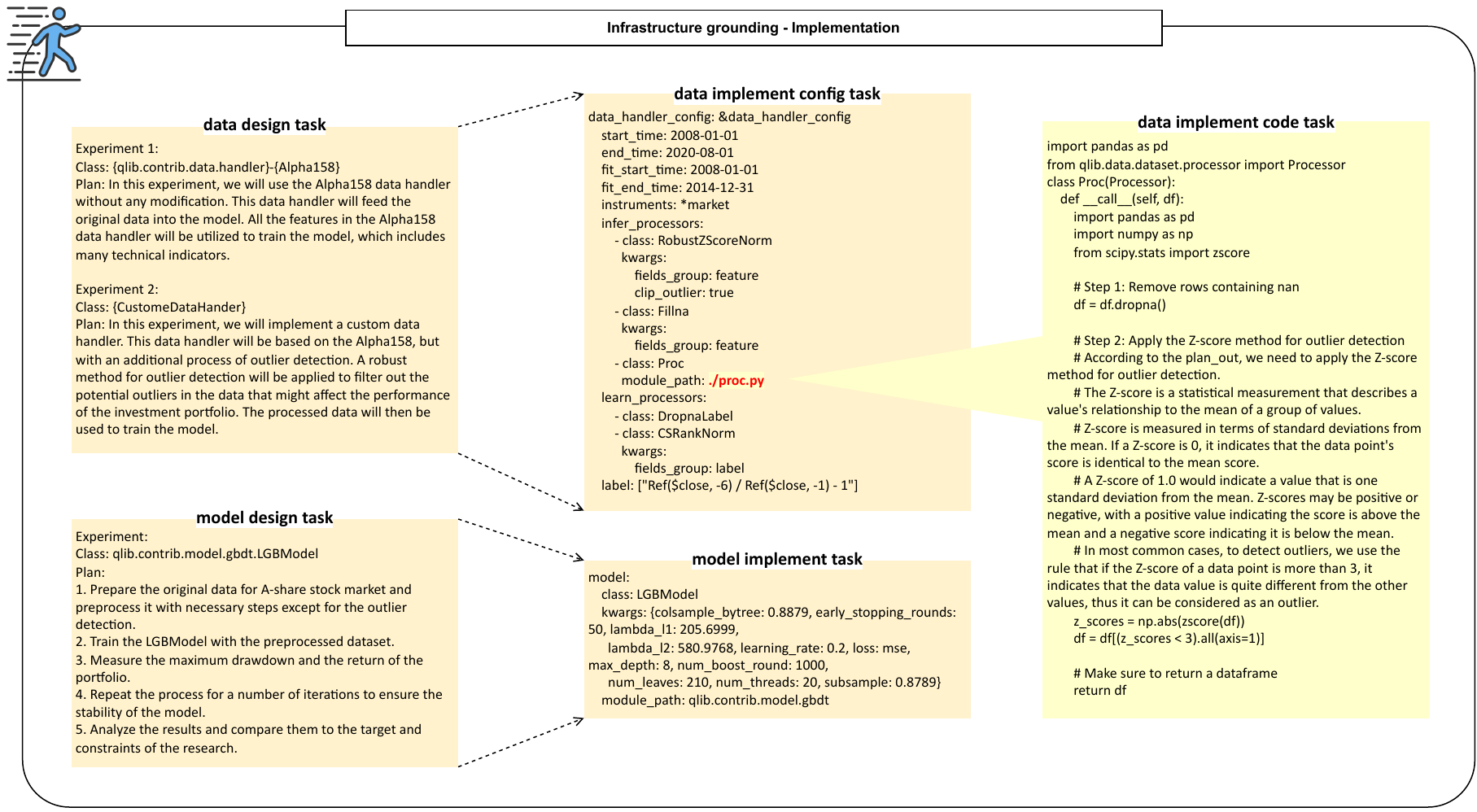}
\caption{Case study on infrastructure grounding. Each module independently implements the detailed plan to runnable configuration and codes. This case shows the detail of an outlier clip experiment implementation.}
\label{fig:case_study_action}
\end{figure} 

To demonstrate how \ppname{} automates R\&D in a more intuitive way, we provide a detailed case in this section. As shown in \ref{fig:formal}, the R\&D loop contains three parts: Requirement analysis, planning \& implementation, experiment execution and knowledge management. Our case will show each part in detail.

Before our case, user has conducted several rounds of experiments on the intention: "I want to build an A-share stock market daily portfolio in quantitative investment and minimize the maximum drawdown while maintaining return." In the former experiments, \ppname{} has done research on missing value, normalization, data splitting and so on.

\paragraph{Requirement analysis.} \ppname{} analyzed the user intention and got the target and constraints in user's intention with the help of domain knowledge.

\paragraph{Planning} As mentioned in \ref{subsection:faaetsl}, \ppname{} plans following the target domain workflow. The planning is conducted by a sequence of sub-tasks. As shown in \ref{fig:case_study_plan}, \ppname{} firstly proposes an idea which considers exploiting former practice knowledge and exploring new direction which has not been touched. Secondly, \ppname{} decides an action to be the target action to do experiments on. About this action, \ppname{} generates a hypothesis and designs a group of experiments to prove or disprove this hypothesis. Finally, \ppname{} points out the difference of each experiments and designs each module's responsibility in the workflow. 

\paragraph{Implementation} The infrastructure grounding process implements the detail plan into runnable projects. As shown in \ref{fig:case_study_action}, \ppname{} feed the detail plan to each module and each module implements a part of the project independently. Take data module as an example. Firstly, \ppname{} chooses the template class for the plan and ground the plan into detailed descriptions from the tool's perspective. Secondly, \ppname{} generates the running configuration on the guidance of the former succeeded knowledge. Finally, \ppname{} writes necessary code to support some personalized function in user's plan. The implementation includes implementing the independent target action which is a processor on dataset or a model hyperparameter and implementing the control actions which should be reasonable and keep the same in all experiments. \ppname{} exploits the practice knowledge and former successful experiments to decide the proper control actions.

\paragraph{Knowledge management} After the experiment execution, \ppname{} generates a practice knowledge based on the execution results. As shown in \ref{fig:KMAll}, The knowledge not only includes the former conclusion but also includes the planning and implementation of the whole workflow. All knowledges are well managed and store for future efficient query.

The detailed case shows a complete loop in data-centric research and development. With the help of LLM, the agent analyzes the user's requirement, decides a research direction, plans the experiments, implements\&executes the experiments and generates useful knowledge automatically. With this autonomous loop, the R\&D process will be remarkably accelerated.

%% file: content/06limdis.tex
\section{Limitations and Discussions}
\label{section:lim}

\paragraph{Challenges in Complex Software Design.}

Based on our experience, the current LLM is capable of writing code within a decoupled module when provided with clear instructions. For instance, when working on data processing, there is a relatively clear interface based on pandas along with instructions.
However, for more complex examples that may be coupled with other parts, LLM may have a relatively higher failure rate.
For instance, when implementing a model, it may be necessary to align the interface with various tools such as sklearn\cite{pedregosa2011scikit}, Pytorch\cite{paszke2019pytorch}, and TensorFlow\cite{abadi2016tensorflow}. The initialization, training, and prediction methods are often interconnected and should be considered together. Additionally, it is important to implement a design that can be shared both online and offline.
Handling complex software design still remains challenging in our scenario.

\paragraph{Co-evolving Between the Infrastructure and R\&D.}
The requirements to fast iterate R\&D incentivize the creation of reusable and advanced infrastructure. An advanced infrastructure facilitates more advanced R\&D.
In daily work life, infrastructure and R\&D evolve together towards advancement.
In this paper, we aim to automate the R\&D evolving cycle. However, we do not consider the evolution of the infrastructure in this study. To the best of our knowledge, this research topic remains unexplored.

%% file: content/07conclu.tex
\section{Conclusion and Future work}
\label{section:concl}

In this paper, we have proposed the use of large language models (LLMs) to assist in R\&D activities. We have focused on data-centric R\&D scenarios and have hypothesized that LLMs can help reduce human effort and enhance the quality of R\&D outputs by providing automated support for idea generation, experiment planning, result analysis, and idea evolution. To test our hypothesis, we have conducted experiments on a concrete industrial scenario in quantitative investment, using Qlib as the domain-specific professional tool. Our results have demonstrated the effectiveness of our approach.

We have identified two major challenges in applying LLMs to automate R\&D: long-horizon planning and specific knowledge requirements. To address these challenges, we have proposed a methodology called "Framework as an Extensible Task-Dependent Symbolic Language" for robust and reliable planning, and have developed a dedicated knowledge management system that supports specific R\&D queries related to ideas, implementation, and tools.
Our work has several contributions. We are the first to formally define the R\&D evolving cycle and apply LLMs to automate industrial R\&D. We have identified the main challenges in applying LLMs to automate R\&D and have proposed solutions to address these challenges. We have conducted extensive experiments on a concrete industrial scenario and have demonstrated the effectiveness of our approach.

In future work, we plan to extend our approach to other data-centric R\&D scenarios and to further improve the performance of LLMs in assisting R\&D activities. We also plan to investigate the integration of LLMs with other AI technologies, such as reinforcement learning and knowledge graphs, to further enhance the capabilities of LLMs in R\&D.
The limitations mentioned in Section \ref{section:lim} can also serve as potential future research directions.

%% file: content/09suppliment.tex
\section{Appendix}
\label{section:appendix}


\subsection{Detailed design of the evaluation tasks and results}
\label{appdix_detail}
In this section, we provide a detailed description of all tasks and offer more detailed results and evaluations of the methods.

\subsubsection{Understanding}
\label{appdix_detail_understanding}

  \paragraph{Task U1.}
    I want to build an A-share stock market daily portfolio in quantitative investment and minimize the maximum drawdown while maintaining return.
    
  \paragraph{Task U2.}
   I want to build an A-share stock market daily portfolio in quantitative investment and maximize the return.

  \paragraph{Task U3.}
   I want to get a portfolio on US stock market which has the highest excess return.

  \paragraph{Task U4.}
   I want to get a portfolio on US stock market which has the highest excess return and smallest maximum drawdown.

  \paragraph{Task U5.}
   Get an A-share stock market daily portfolio with MLP model to maximize the excess return.

  \paragraph{Task U6.}
   I want to get a portfolio on US stock market which has the highest excess return but I don't have a GPU in my computer

  \paragraph{Task U7.}
   I want to get a portfolio on US stock market which has the highest excess return but I only have a GPU with a very small GPU memory

  \paragraph{Task U8.}
   I want to get a portfolio on US stock market which has the highest excess return. I have a very powerful GPU which I want to make the best us
  e of it.

  \paragraph{Task U9.}
   I want earn money from A-share stock market. I think catching the momentum in data is a good way to make money.

  \paragraph{Task U10.}
   I want earn money from A-share stock market. I don't want to use any uninterpretable model.

\begin{table}
    \centering
    
\resizebox{\columnwidth}{!}{
    \begin{tabular}{|c|c|c|c|c|} \hline  
         &  \multicolumn{2}{|c|}{\ppname{} / \ppname{} wo KM }&  \multicolumn{2}{|c|}{Standard / COT}\\ \hline  
 & Target alignment& Constraint alignment& Target alignment& Constraint alignment\\ \hline  
 Task U1&  1&  1&  1&  1\\ \hline  
 Task U2& 1& 1& 1& 1\\ \hline  
 Task U3&  1&  1&  1&  0.8\\ \hline  
 Task U4& 1& 1& 1& 1\\ \hline 
 Task U5&  1&  1&  1&  0.7\\ \hline  
 Task U6& 1& 1& 1& 0.7\\ \hline  
 Task U7&  1&  1&  1&  0.7\\ \hline  
 Task U8& 1& 1& 1& 1\\ \hline  
 Task U9& 1& 1& 1& 1\\ \hline  
 Task U10& 1& 0.8& 1& 0.8\\ \hline  
 Average & \multicolumn{2}{|c|}{0.99}& \multicolumn{2}{|c|}{0.93}\\ \hline
    \end{tabular}
}

    \caption{The overall score of understanding tasks.}
    \label{tab:understand_score}
\end{table}

\begin{figure}
    \centering
    \includegraphics[width=1\linewidth]{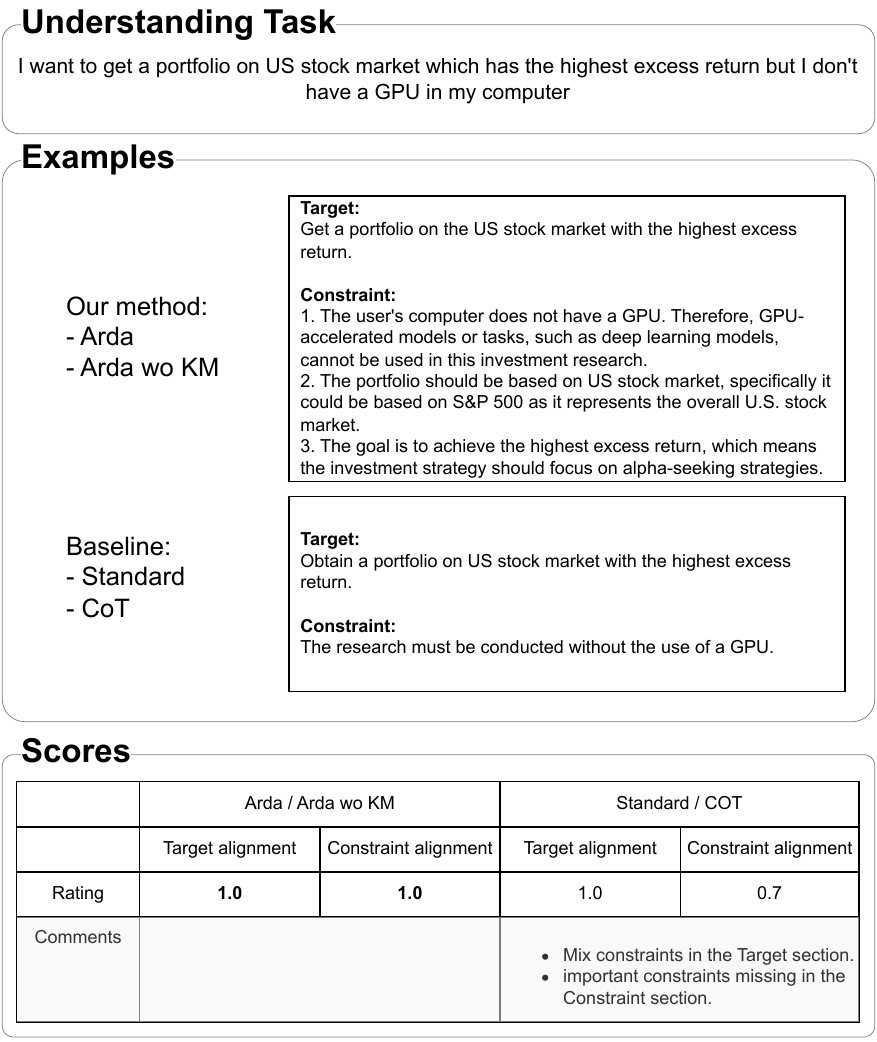}
    \caption{An example of understanding tasks.}
    \label{fig:taskU}
\end{figure}

The scores for all tasks are listed in the Table \ref{tab:understand_score}.  To more intuitively compare different methods, we show a detailed case of Task \ref{tab:understand_score} in Figure \ref{fig:taskU}.

\subsubsection{Exploration \& Exploitation}
\label{appendix_ee}
Exploration \& Exploitation includes two parts. One from cold start and another from warm start. 

The cold start part shares the same tasks as Understanding. The tasks are shown in Section \ref{appdix_detail_understanding}.
The scores are shown in table \ref{tab:coldstart_ee_score}.

The warm start part chooses 5 tasks from Section \ref{appdix_detail_understanding}.
The chose tasks are as listed and the scores are shown in table \ref{tab:warmstart_ee_score}.

    \paragraph{Task E1.}
    I want to build an A-share stock market daily portfolio in quantitative investment and maximize the return.
    
  \paragraph{Task E2.}
   I want to get a portfolio on A-share stock market which has the highest excess return.

  \paragraph{Task E3.}
   Get an A-share stock market daily portfolio with MLP model to maximize the excess return.

  \paragraph{Task E4.}
   I want to get a portfolio on A-share stock market which has the highest excess return but I don't have a GPU in my computer

  \paragraph{Task E5.}
   I want earn money from A-share stock market. I don't want to use any uninterpretable model.

\begin{table}
    \centering
    
\resizebox{\columnwidth}{!}{
    \begin{tabular}{|c|c|c|c|c|c|c|c|c|c|} \hline
    &  \multicolumn{3}{|c|}{Standard} &  \multicolumn{3}{|c|}{COT} &  \multicolumn{3}{|c|}{\ppname{} / \ppname{} wo KM}\\ \hline  
 & Accuracy & Professionalism & Feasibility & Accuracy & Professionalism & Feasibility & Accuracy & Professionalism & Feasibility \\ \hline  
Task U1 & 0.9 & 0.7 & 0.8 & 0.7 & 0.8 & 0.6 & 1.0 & 0.9 & 1.0 \\ \hline  
Task U2 & 0.9 & 0.7 & 0.8 & 0.7 & 0.8 & 0.7 & 0.7 & 0.8 & 0.7 \\ \hline  
Task U3 & 0.9 & 0.7 & 0.8 & 0.7 & 0.8 & 0.7 & 1.0 & 0.9 & 1.0 \\ \hline  
Task U4 & 0.9 & 0.5 & 0.7 & 0.7 & 0.7 & 0.5 & 1.0 & 0.9 & 1.0 \\ \hline  
Task U5 & 0.9 & 0.5 & 0.7 & 0.9 & 0.7 & 0.6 & 1.0 & 0.7 & 1.0 \\ \hline  
Task U6 & 0.9 & 0.7 & 0.8 & 0.9 & 0.8 & 0.7 & 0.8 & 0.9 & 1.0 \\ \hline  
Task U7 & 0.9 & 0.6 & 0.5 & 0.9 & 0.8 & 0.6 & 1.0 & 0.9 & 1.0 \\ \hline  
Task U8 & 0.9 & 0.7 & 0.8 & 0.9 & 0.8 & 0.8 & 0.7 & 0.7 & 1.0 \\ \hline  
Task U9 & 0.7 & 0.6 & 0.4 & 0.8 & 0.7 & 0.7 & 1.0 & 1.0 & 1.0 \\ \hline  
Task U10 & 0.9 & 0.7 & 0.5 & 0.8 & 0.8 & 0.6 & 1.0 & 0.9 & 1.0 \\ \hline  
 Average & \multicolumn{3}{|c|}{0.73}& \multicolumn{3}{|c|}{0.74}& \multicolumn{3}{|c|}{0.917}\\ \hline
\end{tabular}
}
    \caption{The overall score of exploitation \& exploration tasks at cold start scenario. Tasks in this table is the same in understanding.}
    \label{tab:coldstart_ee_score}
\end{table}

\begin{table}
    \centering
    
\resizebox{\columnwidth}{!}{
    \begin{tabular}{|c|c|c|c|c|c|c|c|c|c|c|c|c|} \hline
    &  \multicolumn{3}{|c|}{Standard} &  \multicolumn{3}{|c|}{COT} &  \multicolumn{3}{|c|}{\ppname{}} &  \multicolumn{3}{|c|}{\ppname{} wo KM}\\ \hline  
 & Accuracy & Professionalism & Feasibility & Accuracy & Professionalism & Feasibility & Accuracy & Professionalism & Feasibility & Accuracy & Professionalism & Feasibility \\ \hline  
Task E1 & 0.8 & 0.8 & 0.7 & 0.7 & 0.7 & 0.5 & 0.8 & 0.8 & 0.9 & 0.9 & 0.9 & 0.9 \\ \hline  
Task E2 & 0.8 & 0.8 & 0.5 & 0.7 & 0.8 & 0.4 & 0.8 & 0.8 & 0.9 & 0.9 & 0.8 & 0.9 \\ \hline  
Task E3 & 0.8 & 0.6 & 0.5 & 0.8 & 0.8 & 0.4 & 0.9 & 0.8 & 0.9 & 0.9 & 0.7 & 0.9 \\ \hline  
Task E4 & 0.6 & 0.6 & 0.5 & 0.7 & 0.8 & 0.5 & 0.7 & 0.8 & 0.8 & 0.9 & 0.7 & 0.9 \\ \hline  
Task E5 & 0.7 & 0.6 & 0.5 & 0.7 & 0.8 & 0.5 & 0.8 & 0.8 & 0.9 & 0.9 & 0.9 & 0.9 \\ \hline  
 Weighted average & \multicolumn{3}{|c|}{0.66}& \multicolumn{3}{|c|}{0.685}& \multicolumn{3}{|c|}{0.82}& \multicolumn{3}{|c|}{0.85}\\ \hline
\end{tabular}
}
    \caption{The overall score of exploitation \& exploration tasks at warm start scenario. The exploration is done after 5 round of R\&D loop. The weight is set 0.25, 0.5, 0.25 to Accuracy, Professionalism and Feasibility.}
    \label{tab:warmstart_ee_score}
\end{table}

\subsubsection{Grounding}

\paragraph{Task G1.} 
\begin{itemize}  
    \item DataHandler Plan:  
    \begin{enumerate}  
        \item Use the Alpha158 module to handle data.  
        \item Apply the quant investment strategy on the A-share stock market.  
        \item Do not apply MinMaxNorm normalization on the data.  
    \end{enumerate}
    \item Model Plan:  
    \begin{enumerate}  
        \item Initialize the LGBModel with default parameters. The model will be trained on the normalized data set using the LightGBM gradient boosting framework. The trained model will then be used to predict the future returns of the A-share stock market. The portfolio will be constructed based on these predictions. The maximum drawdown and return of the portfolio will be calculated and compared. The process will be repeated for all the stock trading days in the test period.   
    \end{enumerate}  
\end{itemize}  

\paragraph{Task G2.} 
\begin{itemize}  
    \item DataHandler Plan:  
    \begin{enumerate}  
        \item Use the Alpha158 module to handle data.  
        \item Apply the quant investment strategy on the A-share stock market.  
        \item Apply MinMaxNorm normalization on the data.  
    \end{enumerate}
    \item Model Plan:  
    \begin{enumerate}  
        \item Initialize the LGBModel with default parameters. The model will be trained on the normalized data set using the LightGBM gradient boosting framework. The trained model will then be used to predict the future returns of the A-share stock market. The portfolio will be constructed based on these predictions. The maximum drawdown and return of the portfolio will be calculated and compared. The process will be repeated for all the stock trading days in the test period.   
    \end{enumerate}  
\end{itemize} 

\paragraph{Task G3.} 
\begin{itemize}  
    \item DataHandler Plan:  
    \begin{enumerate}  
        \item Build the DataHandler without applying CSRankNorm normalization to the data.  
    \end{enumerate}
    \item Model Plan:  
   \begin{enumerate}  
    \item Initialize the LGBModel.  
    \item Set the hyperparameters of the model. For instance, set num\_leaves to 31, max\_depth to -1, learning\_rate to 0.05, n\_estimators to 100, subsample\_for\_bin to 200000, objective to 'regression', class\_weight to None, min\_split\_gain to 0.0, min\_child\_weight to 0.001, min\_child\_samples to 20, subsample to 1.0, subsample\_freq to 0, colsample\_bytree to 1.0, reg\_alpha to 0.0, and reg\_lambda to 0.0. Set silent to True and importance\_type to 'split'.  
    \item Train the model with the training dataset.  
    \item Use the trained model to predict the target variable in the test dataset.  
    \item Evaluate the model performance by comparing the predicted values with the actual values.  
    \item Apply the CSRankNorm normalization to the data.  
    \item Run the model again with the normalized data and compare the results with the original model.  
    \item Analyze the difference in model performance with and without the normalization applied.  
    \item Use the model with better performance to manage the investment portfolio on a daily basis.  
    \item Monitor and adjust the model performance over time.  
\end{enumerate}  
  
\end{itemize} 

\paragraph{Task G4.} 
\begin{itemize}  
    \item DataHandler Plan:  
    \begin{enumerate}  
        \item Build the DataHandler with applying CSRankNorm normalization to the data.  
    \end{enumerate}
    \item Model Plan:  
    \begin{enumerate}  
    \item Initialize the LGBModel.  
    \item Set the hyperparameters of the model. For instance, set num\_leaves to 31, max\_depth to -1, learning\_rate to 0.05, n\_estimators to 100, subsample\_for\_bin to 200000, objective to 'regression', class\_weight to None, min\_split\_gain to 0.0, min\_child\_weight to 0.001, min\_child\_samples to 20, subsample to 1.0, subsample\_freq to 0, colsample\_bytree to 1.0, reg\_alpha to 0.0, and reg\_lambda to 0.0. Set silent to True and importance\_type to 'split'.  
    \item Train the model with the training dataset.  
    \item Use the trained model to predict the target variable in the test dataset.  
    \item Evaluate the model performance by comparing the predicted values with the actual values.  
    \item Apply the CSRankNorm normalization to the data.  
    \item Run the model again with the normalized data and compare the results with the original model.  
    \item Analyze the difference in model performance with and without the normalization applied.  
    \item Use the model with better performance to manage the investment portfolio on a daily basis.  
    \item Monitor and adjust the model performance over time.  
\end{enumerate}  
\end{itemize}

\paragraph{Task G5.}
\begin{itemize}
\item DataHandler Plan:
\begin{enumerate}
\item Apply the Alpha158 DataHandler to the original data without any normalization process. This will serve as the baseline experiment.
\end{enumerate}
\item Model Plan:
\begin{enumerate}
\item Initialize the LSTM model by creating an instance of the class. Use the default hyperparameters of the LSTM model for the experiment.
\item The data will be fed to the model without any preprocessing in one experiment and with MinMaxNorm normalization in the other experiment.
\item Remember to set the same random seed to ensure the experiments are comparable.
\end{enumerate}
\end{itemize}

\paragraph{Task G6.}
\begin{itemize}
\item DataHandler Plan:
\begin{enumerate}
\item Apply the Alpha158 DataHandler to the data with MinMaxNorm normalization. This modification is expected to improve the excess return of the portfolio. The MinMaxNorm normalization will be applied to the data before feeding it into the DataHandler.
\end{enumerate}
\item Model Plan:
\begin{enumerate}
\item Initialize the LSTM model by creating an instance of the class. Use the default hyperparameters of the LSTM model for the experiment.
\item The data will be fed to the model without any preprocessing in one experiment and with MinMaxNorm normalization in the other experiment.
\item Remember to set the same random seed to ensure the experiments are comparable.
\end{enumerate}
\end{itemize}

\paragraph{Task G7.}   
\begin{itemize}    
    \item DataHandler Plan:    
    \begin{enumerate}    
        \item Apply ZScoreNorm preprocessing to the data.    
        \item Use the MLP model for portfolio construction.    
        \item Update the portfolio daily.    
        \item Measure the excess return of the portfolio.    
    \end{enumerate}  
    \item Model Plan:    
    \begin{enumerate}    
        \item Initialize the DNNModelPytorch (MLP model in Pytorch) in Qlib.    
        \item Set up the model with the appropriate parameters.    
        \item Train the model with the daily A-share stock market data.    
        \item Use the trained model to predict the daily portfolio.    
        \item Calculate the excess return of the portfolio.    
        \item Record the model's performance.    
        \item Repeat steps 3-6 for each day.    
    \end{enumerate}    
\end{itemize}

\paragraph{Task G8.}   
\begin{itemize}    
    \item DataHandler Plan:    
    \begin{enumerate}    
        \item Use the raw data without preprocessing.    
        \item Use the MLP model for portfolio construction.    
        \item Update the portfolio daily.    
        \item Measure the excess return of the portfolio.    
    \end{enumerate}  
    \item Model Plan:    
    \begin{enumerate}    
        \item Initialize the DNNModelPytorch (MLP model in Pytorch) in Qlib.    
        \item Set up the model with the appropriate parameters.    
        \item Train the model with the daily A-share stock market data.    
        \item Use the trained model to predict the daily portfolio.    
        \item Calculate the excess return of the portfolio.    
        \item Record the model's performance.    
        \item Repeat steps 3-6 for each day.    
    \end{enumerate}    
\end{itemize}

\paragraph{Task G9.}   
\begin{itemize}    
    \item DataHandler Plan:    
    \begin{enumerate}    
        \item Initialize the Alpha158 DataHandler class.    
        \item Load the US stock market data.    
        \item Normalize the data if required.    
        \item Set the model to use the Alpha158 DataHandler for handling the data.    
        \item Monitor the excess return and training time as metrics.    
        \item Apply early stopping during model training.    
        \item Experiment with the model, recording the results.    
    \end{enumerate}  
    \item Model Plan:    
    \begin{enumerate}    
        \item Initialize the LGBModel with default parameters.    
        \item Set the `early\_stopping` parameter to False.    
        \item Train the model using the given data.    
        \item Evaluate the model's excess return and training time.    
    \end{enumerate}    
\end{itemize}

\paragraph{Task G10.}   
\begin{itemize}    
    \item DataHandler Plan:    
    \begin{enumerate}    
        \item Initialize the Alpha158 DataHandler class.    
        \item Load the US stock market data.    
        \item Normalize the data if required.    
        \item Set the model to use the Alpha158 DataHandler for handling the data.    
        \item Monitor the excess return and training time as metrics.    
        \item Apply early stopping during model training.    
        \item Experiment with the model, recording the results.    
    \end{enumerate}  
    \item Model Plan:    
    \begin{enumerate}    
        \item Initialize the LGBModel with default parameters.    
        \item Set the `early\_stopping` parameter to True.    
        \item Train the model using the given data.    
        \item Evaluate the model's excess return and training time.    
    \end{enumerate}    
\end{itemize}

\paragraph{Task G11.}
\begin{itemize}
\item DataHandler Plan: In this experiment, we will utilize the Alpha158 DataHandler without applying the RobustZScoreNorm in data pre-processing. We will try to construct and optimize the portfolio with the maximum possible excess return on the U.S. stock market. The GPU's limited memory will be taken into account.
\item Model Plan:
\begin{enumerate}
\item Initialize the Tabnet model with parameters suitable for small GPU memory. The parameters are as follows: \newline
- n\_d: 8, the dimension of the prediction layer (usually between 4 to 64). \newline
- n\_a: 8, the dimension of the attention layer (usually between 4 to 64). \newline
- n\_steps: 3, the total number of steps in the decision steps (usually between 1 to 10). \newline
- gamma: 1.3, the coefficient for feature reusage in the decision steps. \newline
- n\_independent: 2, number of independent GLU layer in each GLU block. \newline
- n\_shared: 2, number of shared GLU layer at the first of each decision step. \newline
- epsilon: 1e-15, the stability value to avoid division by zero. \newline
- virtual\_batch\_size: 128, the size of the mini batches for each step. \newline
- momentum: 0.02, the value for momentum in batch normalization. \newline
- mask\_type: "sparsemax", the type of masking function to use.
\item Train the model on the U.S. stock market data with RobustZScoreNorm applied.
\item Evaluate the model on the validation data to monitor the training progress.
\item After the training is complete, use the model to construct a portfolio and calculate the excess return.
\end{enumerate}
\end{itemize}

\paragraph{Task G12.}
\begin{itemize}
\item DataHandler Plan: In this experiment, we will implement the RobustZScoreNorm in the data pre-processing step with the Alpha158 DataHandler. We will then construct and optimize the portfolio focusing on the U.S. stock market while still considering the limited memory of the GPU. We aim to compare the performance of this experiment with Experiment 1 in terms of achieving the highest possible excess return.
\item Model Plan:
\begin{enumerate}
\item Initialize the Tabnet model with parameters suitable for small GPU memory. The parameters are as follows: \newline
- n\_d: 8, the dimension of the prediction layer (usually between 4 to 64). \newline
- n\_a: 8, the dimension of the attention layer (usually between 4 to 64). \newline
- n\_steps: 3, the total number of steps in the decision steps (usually between 1 to 10). \newline
- gamma: 1.3, the coefficient for feature reusage in the decision steps. \newline
- n\_independent: 2, number of independent GLU layer in each GLU block. \newline
- n\_shared: 2, number of shared GLU layer at the first of each decision step. \newline
- epsilon: 1e-15, the stability value to avoid division by zero. \newline
- virtual\_batch\_size: 128, the size of the mini batches for each step. \newline
- momentum: 0.02, the value for momentum in batch normalization. \newline
- mask\_type: "sparsemax", the type of masking function to use.
\item Train the model on the U.S. stock market data with RobustZScoreNorm applied.
\item Evaluate the model on the validation data to monitor the training progress.
\item After the training is complete, use the model to construct a portfolio and calculate the excess return.
\end{enumerate}
\end{itemize}

\paragraph{Task G13.}   
\begin{itemize}    
    \item DataHandler Plan:    
    \begin{enumerate}    
        \item Normalize the data using RobustZScoreNorm technique.    
        \item Utilize the GPU capabilities for data processing and model training.    
        \item The scope of the data should cover the US stock market.    
    \end{enumerate}  
    \item Model Plan:    
    \begin{enumerate}    
        \item Initialize the Transformer model with the given hyperparameters. Since the model is a control variable and not the focus of the experiment, we will use the default hyperparameters provided by Qlib for the transformer model. The process of initializing the model will be the same in all experiments.  
        \item As the experiment is designed to fully utilize the user's powerful GPU, ensure that the GPU is used for model training and inference.    
    \end{enumerate}    
\end{itemize}

\paragraph{Task G14.}   
\begin{itemize}    
    \item DataHandler Plan:    
    \begin{enumerate}    
        \item Normalize the data using other normalization techniques.    
        \item Utilize the GPU capabilities for data processing and model training.    
        \item The scope of the data should cover the US stock market.    
    \end{enumerate}  
    \item Model Plan:    
    \begin{enumerate}    
        \item Initialize the Transformer model with the given hyperparameters. Since the model is a control variable and not the focus of the experiment, we will use the default hyperparameters provided by Qlib for the transformer model. The process of initializing the model will be the same in all experiments.  
        \item As the experiment is designed to fully utilize the user's powerful GPU, ensure that the GPU is used for model training and inference.    
    \end{enumerate}    
\end{itemize}

\paragraph{Task G15.}
\begin{itemize}
\item DataHandler Plan: Use the Alpha158 DataHandler to generate factor-based features for the A-share stock market. This will support capturing the momentum trends in the data and help generate the factor-based models for the investment strategy.
\item Model Plan: Initialize the IGMTF model, which utilizes the factor-based structure to capture momentum trends in the data. Set the hyperparameters as per the requirements of the Qlib IGMTF model.
\end{itemize}

\paragraph{Task G16.}
\begin{itemize}
\item DataHandler Plan: Use the Alpha158 DataHandler to generate factor-based features for the A-share stock market. This will support capturing the momentum trends in the data and help generate the factor-based models for the investment strategy.
\item Model Plan: Initialize the LSTM model, which does not use a factor-based structure. Set the hyperparameters as per the requirements of the Qlib LSTM model.
\end{itemize}

\paragraph{Task G17.}   
\begin{itemize}    
    \item DataHandler Plan:   
    \begin{enumerate}  
        \item Use the Alpha158 DataHandler module.  
        \item Apply RobustZScoreNorm for data pre-processing.  
        \item Build and test the model with the pre-processed data.  
        \item Compare the model's performance with the second experiment.  
    \end{enumerate}  
    \item Model Plan:   
    \begin{enumerate}  
        \item Initialize the LGBModel in Qlib.  
        \item Train the model with the training dataset.  
        \item Evaluate the model's performance with the validation dataset.  
        \item Apply the model to the testing dataset to generate the investment solution.  
        \item Record the performance of the investment solution.  
        \item Repeat steps 2-5 for several rounds to verify the robustness of the solution.  
        \item This process should be done under the same data pre-processing method, which is RobustZScoreNorm in this case. The same RobustZScoreNorm setting should be used across all experiments to ensure the results are comparable.  
    \end{enumerate}  
\end{itemize}  
   
\paragraph{Task G18.}   
\begin{itemize}    
    \item DataHandler Plan:   
    \begin{enumerate}  
        \item Use the Alpha158 DataHandler module.  
        \item Do not apply RobustZScoreNorm for data pre-processing.  
        \item Build and test the model with the non-preprocessed data.  
        \item Compare the model's performance with the first experiment.  
    \end{enumerate}  
    \item Model Plan:   
    \begin{enumerate}  
        \item Initialize the LGBModel in Qlib.  
        \item Train the model with the training dataset.  
        \item Evaluate the model's performance with the validation dataset.  
        \item Apply the model to the testing dataset to generate the investment solution.  
        \item Record the performance of the investment solution.  
        \item Repeat steps 2-5 for several rounds to verify the robustness of the solution.  
        \item This process should be done under the same data pre-processing method, which is RobustZScoreNorm in this case. The same RobustZScoreNorm setting should be used across all experiments to ensure the results are comparable.  
    \end{enumerate}  
\end{itemize}

\paragraph{Task G19.}   
\begin{itemize}    
    \item DataHandler Plan: Apply MinMaxNorm normalization method to the US stock market data and assess the model performance in terms of excess return and maximum drawdown.  
    \item Model Plan: Initialize the LSTM model with default parameters. The model will be trained using the US stock market data without applying any normalization methods. As the model itself is a control variable, the plan only involves initializing the model, not tuning any hyperparameters.  
\end{itemize}  
   
\paragraph{Task G20.}   
\begin{itemize}    
    \item DataHandler Plan: Apply ZScoreNorm normalization method to the US stock market data and assess the model performance in terms of excess return and maximum drawdown.  
    \item Model Plan: Initialize the LSTM model with default parameters. The model will be trained using the US stock market data without applying any normalization methods. As the model itself is a control variable, the plan only involves initializing the model, not tuning any hyperparameters.  
\end{itemize}

\begin{table}  
    \centering  
  
\resizebox{\columnwidth}{!}{%
    \begin{tabular}{|c|c|c|c|c|c|c|} \hline    
         &  \multicolumn{2}{|c|}{\ppname{}}&  \multicolumn{2}{|c|}{Standard}&  \multicolumn{2}{|c|}{CoT}\\ \hline    
         & Pass rate & Planning alignment & Pass rate & Planning alignment & Pass rate & Planning alignment \\ \hline    
         Task G1 & 1 & 1 & 1 & 1 & 1 & 1 \\ \hline
Task G2 & 1 & 1 & 1 & 1 & 1 & 1 \\ \hline
Task G3 & 1 & 1 & 1 & 1 & 1 & 1 \\ \hline
Task G4 & 1 & 1 & 1 & 1 & 1 & 1 \\ \hline
Task G5 & 0 & 0.5 & 1 & 1 & 1 & 1 \\ \hline
Task G6 & 0 & 0.5 & 1 & 1 & 1 & 1 \\ \hline
Task G7 & 1 & 1 & 1 & 1 & 1 & 1 \\ \hline
Task G8 & 1 & 1 & 1 & 1 & 1 & 1 \\ \hline
Task G9 & 1 & 1 & 1 & 1 & 1 & 1 \\ \hline
Task G10 & 1 & 1 & 1 & 1 & 1 & 1 \\ \hline
Task G11 & 1 & 1 & 1 & 1 & 1 & 1 \\ \hline
Task G12 & 1 & 1 & 1 & 1 & 1 & 1 \\ \hline
Task G13 & 0 & 0.8 & 0 & 0.8 & 1 & 1 \\ \hline
Task G14 & 0 & 0.6 & 0 & 0.6 & 1 & 1 \\ \hline
Task G15 & 0 & 0.5 & 0 & 0.7 & 1 & 0.9 \\ \hline
Task G16 & 1 & 1 & 1 & 1 & 1 & 1 \\ \hline
Task G17 & 0 & 0.5 & 0 & 0.5 & 1 & 1 \\ \hline
Task G18 & 0 & 0.5 & 0 & 0.5 & 1 & 1 \\ \hline
Task G19 & 1 & 1 & 1 & 1 & 1 & 1 \\ \hline
Task G20 & 1 & 1 & 1 & 1 & 1 & 1 \\ \hline   
         Overall &  \multicolumn{2}{|c|}{0.747} &  \multicolumn{2}{|c|}{0.825} &  \multicolumn{2}{|c|}{0.995} \\ \hline  
    \end{tabular}%
}  

    \caption{The overall score of grounding tasks}
    \label{tab:groundscore}
\end{table}

\begin{figure}
    \centering
    \includegraphics[width=1\linewidth]{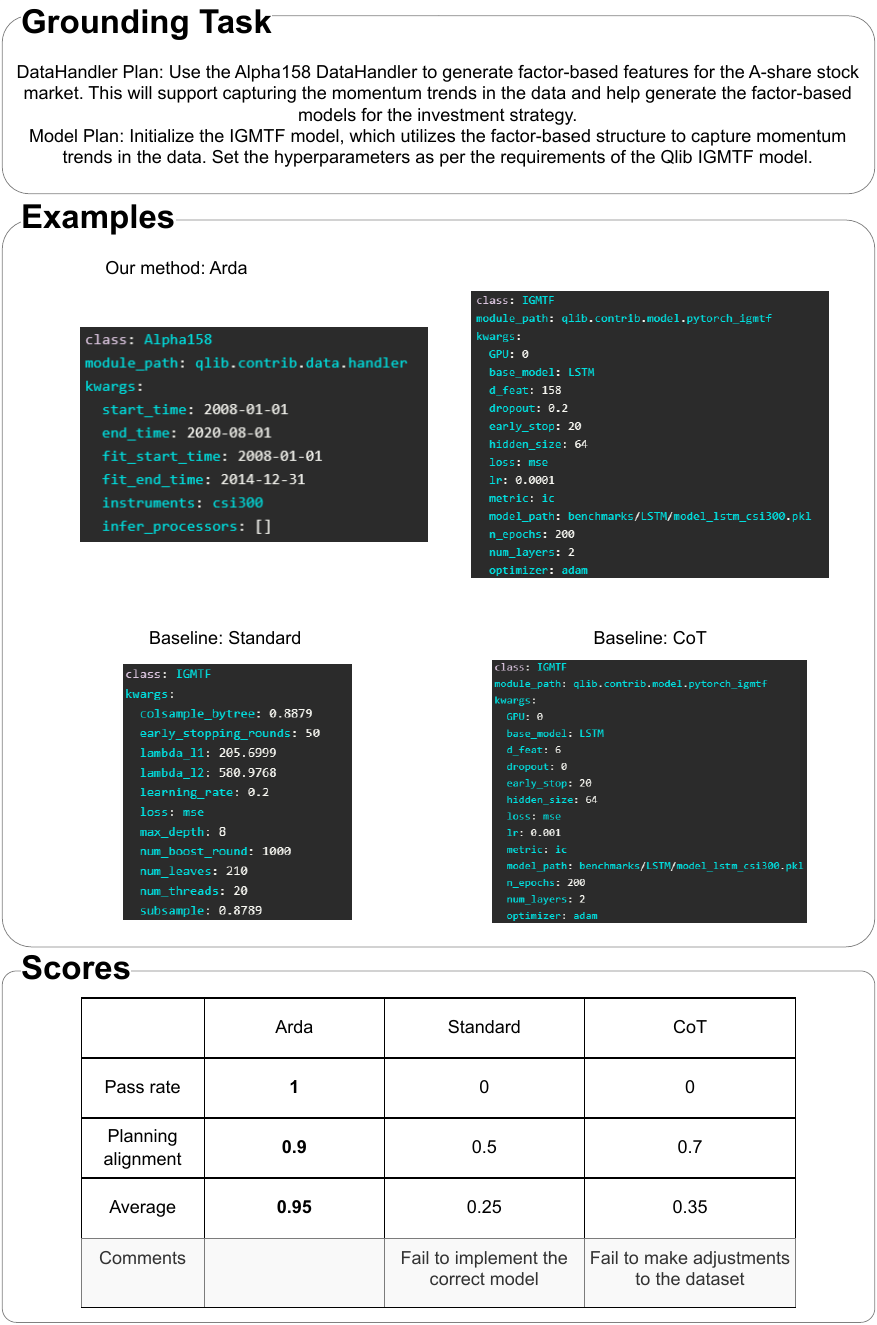}
    \caption{An example of grounding tasks.}
    \label{fig:task-example}
\end{figure}
The scores for all tasks are listed in the Table \ref{tab:groundscore}.  To more intuitively compare different methods, we show a detailed case of Task G15 in Figure \ref{fig:task-example}.

\subsubsection{Transferability}
The transferability injects the knowledge E1 to E5 from tasks in \ref{appendix_ee} and choose another 5 tasks to test the transferability of the system. The difference between T1 to T5 and E1 to E5 is the target. E1 to E5 mainly focuses on return while new tasks focus on several new targets.

New tasks are as follow:

    \paragraph{Task T1.}
    I want to get a portfolio on A-share stock market which has the highest excess return and smallest maximum drawdown.
    
  \paragraph{Task T2.}
   Build an A-share stock market daily portfolio in quantitative investment and minimize the maximum drawdown while maintaining return.

  \paragraph{Task T3.}
   I want to get a portfolio on A-share stock market which has the highest Sharpe ratio.

  \paragraph{Task T4.}
   I want to get a portfolio on A stock market which has the highest annualized return.

  \paragraph{Task T5.}
   I want to get a machine learning model on A stock market which has the highest ICIR.

\begin{table}
    \centering
    
\resizebox{\columnwidth *  2 / 3}{!}{
    \begin{tabular}{|c|c|c|c|c|c|c|c|c|c|c|c|c|} \hline
    &  \multicolumn{2}{|c|}{\ppname{} wo KM} &  \multicolumn{2}{|c|}{\ppname{}}\\ \hline  
 & exploitation & exploration & exploitation & exploration \\ \hline  
Task T1 & 1.0 & 0.9 & 1.0 & 0.9 \\ \hline  
Task T2 & 0.9 & 0.9 & 0.9 & 1.0 \\ \hline  
Task T3 & 0.9 & 0.7 & 1.0 & 0.9 \\ \hline  
Task T4 & 0.8 & 0.8 & 1.0 & 1.0 \\ \hline  
Task T5 & 0.9 & 0.7 & 1.0 & 0.9 \\ \hline  
 average & \multicolumn{2}{|c|}{0.85}& \multicolumn{2}{|c|}{0.96}\\ \hline
\end{tabular}
}
    \caption{The overall score of transferability tasks. }
    \label{tab:appendix_Transferability}
\end{table}

\begin{figure}
    \centering
    \includegraphics[width=1\linewidth]{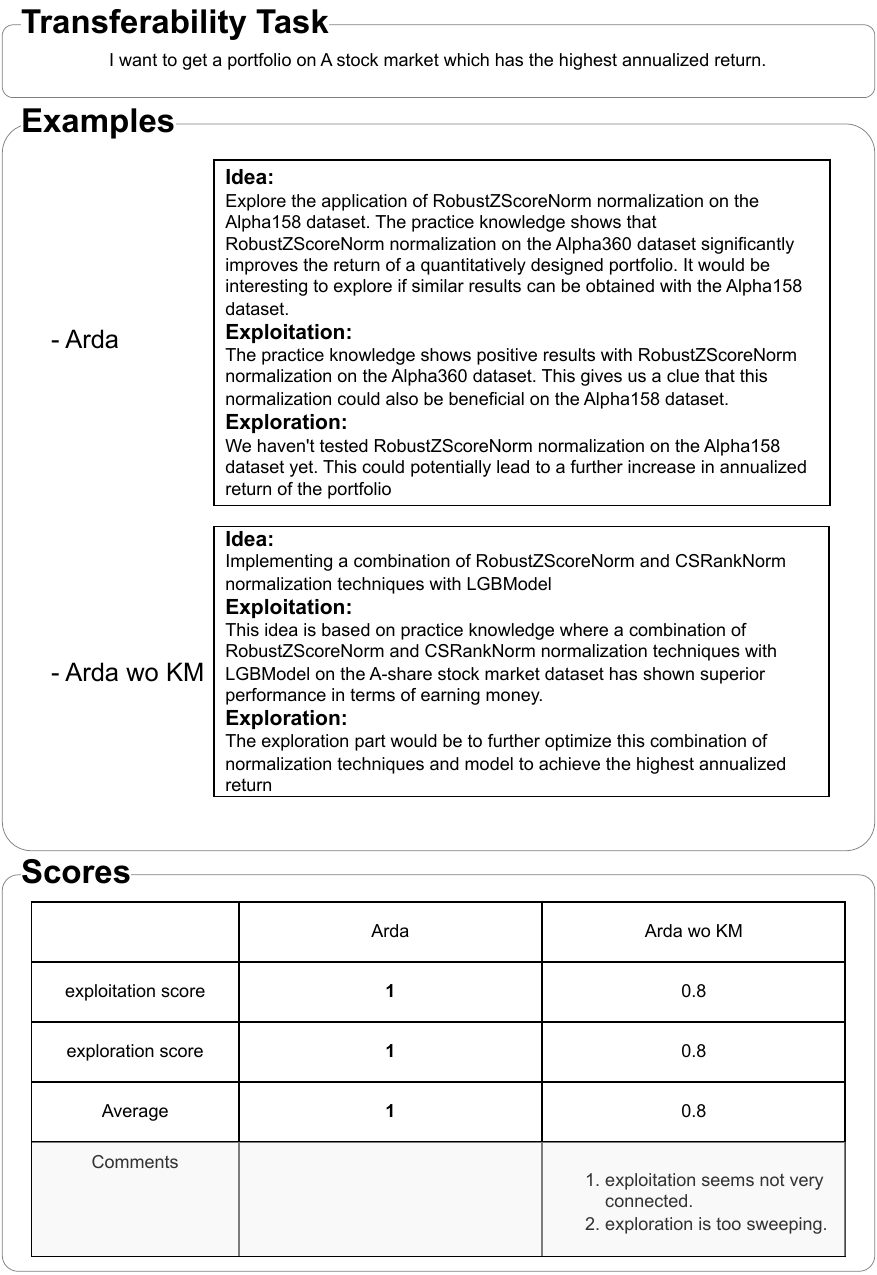}
    \caption{An example of transferability  tasks.}
    \label{fig:tranferability-task-example}
\end{figure}

The results are shown in Table \ref{tab:appendix_Transferability}. A detailed task example is shown in \ref{fig:tranferability-task-example} to further demonstrate the difference between methods.

